\documentclass[sigconf]{acmart}

\usepackage{mathtools, nccmath}
\usepackage{amsmath}
\newcommand{\R}{\mathbb R}
\usepackage{subfigure}
\usepackage{multirow}

\newcommand{\TSCP}{$TS-CP^{2} $}
\AtBeginDocument{%
  \providecommand\BibTeX{{%
    \normalfont B\kern-0.5em{\scshape i\kern-0.25em b}\kern-0.8em\TeX}}}

\setcopyright{acmcopyright}

\copyrightyear{2021}
\acmYear{2021}
\acmConference[WWW '21]{Proceedings of the Web Conference 2021}{April 19--23, 2021}{Ljubljana, Slovenia}
\acmBooktitle{Proceedings of the Web Conference 2021 (WWW '21), April 19--23, 2021, Ljubljana, Slovenia}
\acmPrice{}
\acmDOI{10.1145/3442381.3449903}
\acmISBN{978-1-4503-8312-7/21/04}

\setcopyright{iw3c2w3}
\begin{document}

\title{Time Series Change Point Detection with Self-Supervised Contrastive Predictive Coding}

\author{Shohreh Deldari}
\email{shohreh.deldari@rmit.edu.au}

\affiliation{%
  \institution{School of Computing Technologies, RMIT University}
  \city{Melbourne}
  \state{VIC}
  \country{Australia}
}

\author{Daniel V. Smith}
\email{daniel.v.smith@data61.csiro.au}
\affiliation{%
  \institution{Data61, CSIRO}
  \city{Hobart}
  \state{TAS}
  \country{Australia}
}

\author{Hao Xue}
\email{hao.xue@rmit.edu.au}
\affiliation{%
  \institution{School of Computing Technologies, RMIT University}
  \city{Melbourne}
  \state{VIC}
  \country{Australia}
}

\author{Flora D. Salim}
\email{flora.salim@rmit.edu.au}
\affiliation{%
  \institution{School of Computing Technologies, RMIT University}
  \city{Melbourne}
  \state{VIC}
  \country{Australia}
}
\renewcommand{\shortauthors}{Shohreh Deldari, Daniel V. Smith, Hao Xue, Flora D. Salim}

\begin{abstract} Change Point Detection (CPD) methods identify the times associated with changes in the trends and properties of time series data in order to describe the underlying behaviour of the system. For instance, detecting the changes and anomalies associated with web service usage, application usage or human behaviour can provide valuable insights for downstream modelling tasks.
We propose a novel approach for self-supervised \textbf{T}ime \textbf{S}eries \textbf{C}hange \textbf{P}oint detection method based on \textbf{C}ontrastive \textbf{P}redictive coding (\TSCP ). \TSCP\ is the first approach to employ a contrastive learning strategy for CPD by learning an embedded representation that separates pairs of embeddings of time adjacent intervals from pairs of interval embeddings separated across time. Through extensive experiments on three diverse, widely used time series datasets, we demonstrate that our method outperforms five state-of-the-art CPD methods, which include unsupervised and semi-supervised approaches. \TSCP\ is shown to improve the performance of methods that use either handcrafted statistical or temporal features by 79.4\% and deep learning-based methods by 17.0\% with respect to the F1-score averaged across the three datasets.

\end{abstract}




 \maketitle

\section{Introduction}
\label{sec:intro}

The ubiquity of digital technologies along with the substantial 
processing power and storage capacity on offer means we currently have an unprecedented 
ability to access and analyse data. The scale and  
velocity in which data is
being stored and shared, however, means that we often lack the resources to utilise
traditional data curation processes. For instance, in supervised machine learning approaches, the data annotation process can be an expensive, unwieldy and inaccurate one. Consequently, this is why self-supervised and unsupervised learning methods are currently hot topics in the machine learning community where the goal is to maximise the value of raw data.

Change point detection (CPD), an analytical method to identify the times 
associated with abrupt transitions of a series 
can be used to extract meaning from non-annotated data. Change points, whether they have been generated from video cameras, microphones, environmental sensors or mobile applications can provide a critical understanding
of the underlying behaviour of the system being modelled. For instance, change points can represent alterations in the system state that might require human attention, such as a system fault or an upcoming emergency. Furthermore, CPD methods can be employed in related problems of temporal segmentation, event detection and temporal anomaly detection.

\begin{figure}[htp]
\centering
  \includegraphics[width=\linewidth]{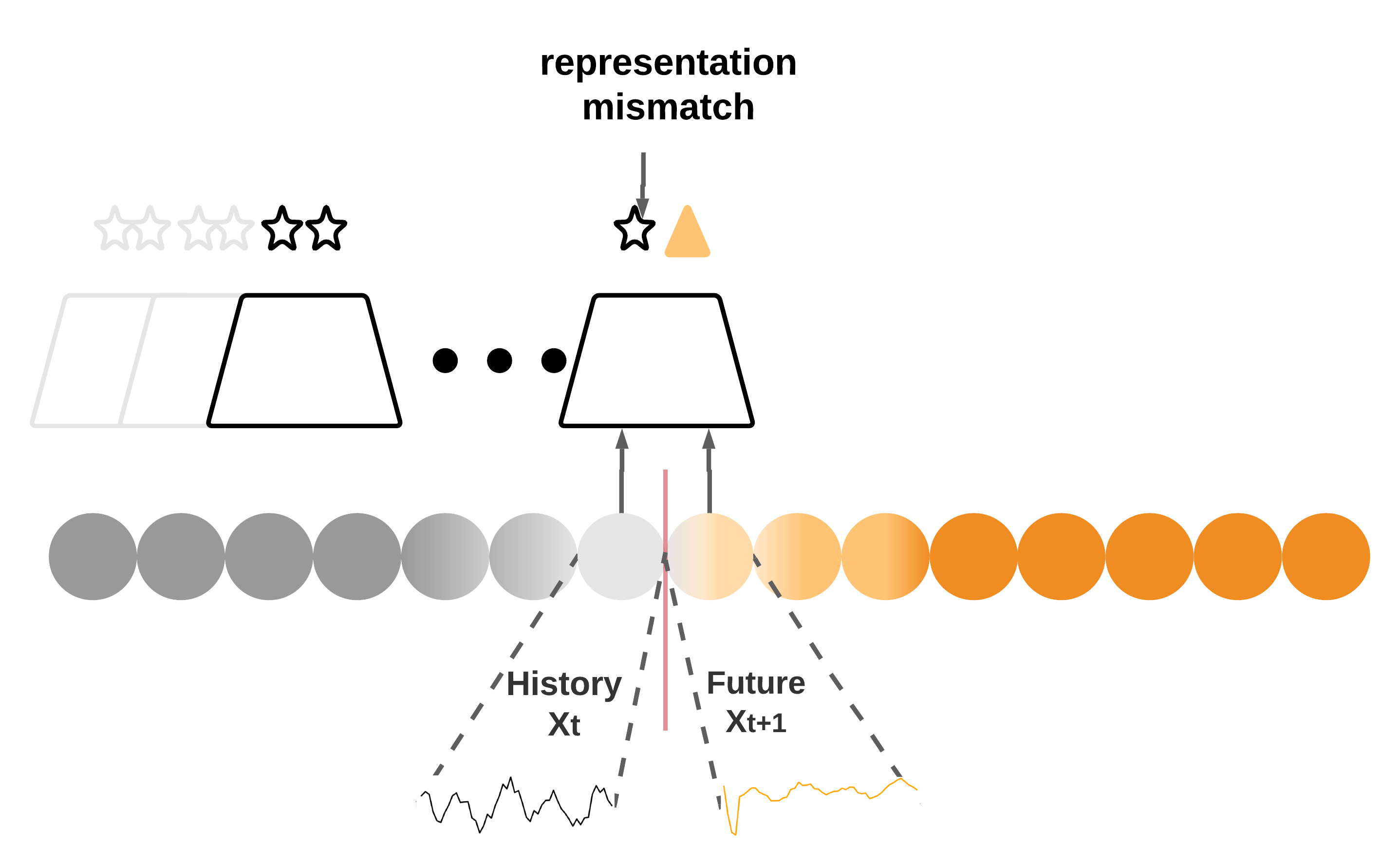}
  \caption{Overview of presented change point detection approach based on predictive representation learning. }

  \label{fig:main_idea}
\end{figure}

CPD techniques have been applied to multivariate time series data in a broad range of research areas including network traffic analysis \cite{kurt2018bayesian}, IoT applications and smart homes \cite{aminikhanghahi2019enhancing}, human activity recognition (HAR) \cite{liono2016optimal,shoaib2016complex, wang2018modeling, chamroukhi2013joint, deldari2020Espresso}, human physiological and emotional analysis \cite{deldari2020Espresso}, factory automation \cite{xia2020robust}, trajectory prediction \cite{Sadri2018tajectory}, user authentication \cite{huang2019id}, life-logging \cite{chavarriaga2013opportunity}, elderly rehabilitation \cite{lam2016automated}, and daily work routine studies \cite{deldari2016inferring}. In addition to time series, CPD can applied to other data modalities with a temporal dimension, such as video, where it has been used for video captioning \cite{ding2018weakly,farha2019ms} and video summarising \cite{aakur2019perceptual,wei2018sequence} applications.

Change points are commonly estimated from one of a number of different properties of a time series, including its temporal continuity, distribution or shape. Unsupervised CPD methods are generally developed to identify changes based upon one particular property. For instance, \textit{FLOSS} \cite{gharghabi2018domain} was developed to detect changes in the temporal shape, whilst \textit{RuLSIF} \cite{liu2013change} and aHSIC \textit{\cite{Yamadaahsic}} were developed to identify changes in the statistical distribution. Current CPD methods have failed to generalise effectively \cite{deldari2020Espresso} as
the semantic boundaries of different applications will usually be associated with different time series properties. For example, abnormalities in the rhythm of the human heart are best characterised by changes in the temporal shape pattern of an electrocardiogram (ECG), whereas changes in human posture (as measured with an RFID sensor system) are best characterised by abrupt statistical changes. In this case, the detection performance degrades when a statistical CPD method is applied to the heart beat application, whilst shape based CPD methods will fail in the human posture application.  Furthermore, for many applications in which data is 
continuously collected, 
time series with be characterised by 
slowly varying temporal shape and 
statistical properties. The change points associated
with such time series can be subtle and remain a 
challenge for CPD methods to address.

In this work, we propose \TSCP ~, a novel approach for self-supervised \textbf{T}ime \textbf{S}eries \textbf{C}hange \textbf{P}oint detection method based on \textbf{C}ontrastive \textbf{P}redictive coding. We pose the question of whether self-supervised learning can be used to provide an effective, general representation for CPD. The intuition here is to exploit the local correlation present within a time series by learning a representation that maximises the shared information between contiguous time intervals, whilst minimising the shared information between pairs of time intervals that are separated in time (i.e. pairs of time intervals with less correlation). It is hypothesised that whenever the learnt representation differs significantly between time adjacent intervals, a change point is more likely to be present. 

We aim to show that this self-supervised representation is capable of detecting a broader range of change points than previous methods that have been specifically designed to exploit a narrow scope of time series properties (i.e. commonly either its temporal continuity, distribution or shape patterns).  Figure \ref{fig:main_idea} shows a high-level overview of the approach, which is the first CPD approach based upon contrastive representation learning. Furthermore, whilst there are contrastive learning methods for image \cite{chen2020simclr,sohn2016improved},  audio \cite{oord2018representation,saeed2020contrastive} and text \cite{oord2018representation}, this is the first approach utilising contrastive learning on general time series, which in turn, introduces some unique challenges. Furthermore, our technique does not rely on any assumptions about the statistical distribution of the data making it applicable to a broad range of real-world applications. The main contributions of our paper are as follows:

\begin{itemize}
    \item We leverage contrastive learning as an unsupervised objective function for the CPD task. To the best of our knowledge, we are the first to employ contrastive learning to the CPD problem.
    
      \item We propose a representation learning framework to tackle the problem of self-supervised CPD by capturing compact, latent embeddings that represent historical and future time intervals of the times series. 
      
      \item We compare our proposed method against five state of the art CPD methods, which include deep learning and non deep learning based methods, investigate the benefits of each through extensive experiments.

      \item We investigate the performance impact of the hyperparameters used within our self-supervised learning method including batch size, code size, and window size.

\end{itemize}

To make \TSCP\ reproducible, all the code, data and experiments are available in the project's web page \footnote{https://github.com/cruiseresearchgroup/TSCP2}.

\section{Related Work and Background}
\label{sec:related}
In this section, we review existing approaches for the CPD problem. Since we employ contrastive learning for our time series change point detection method, we also outline recent works on self-supervised contrastive learning. We will then 
review recent representation learning approaches, 
not only for time series data, but other data modalities as well.

\subsection{Time series change point detection}
Although self-supervised learning methods have recently attracted the interest of the deep learning community, current CPD methods are mostly based on non deep learning approaches yet. 

Existing approaches can be categorised based upon the features of the 
time series that they consider for CPD. Statistical methods often compute change points on the basis of 
identifying statistical differences between adjacent short intervals of a time series. The statistical differences between intervals are usually measured with either parametric or non-parametric approaches. Parametric methods use a Probability Density Function (PDF) such as \cite{Basseville1993cdetection} or auto-regressive model \cite{yamanishi2002unifying} to represent the time intervals, however, such convenient representations limit the types of statistical changes that can be detected.
Non-parametric methods offer a greater degree of flexibility to represent the density functions of time intervals by utilising kernel functions. Estimating the ratio of time interval PDFs is a simpler problem to address
than estimating the individual PDFs of the time intervals. The methods in \textit{RuLSIF} \cite{liu2013change}, \textit{KLIEP} \cite{Kawahara2012Seqeuntial} and \textit{SEP} \cite{aminikhanghahi2019enhancing} used a non-parametric Gaussian kernel
to model the density ratio distribution between subsequent time intervals. 
\cite{Yamadaahsic} detected abrupt change points by 
calculating separability of adjacent intervals based on 
kernel-based additive Hilbert-Schmidt Independence Criterion (\textit{aHSIC}). Kernel approaches assume there is statistical homogeneity within each interval, which can be problematic for change point detection. Furthermore, kernel functions often require parameters to be carefully tuned. 

There is another category of statistical CPD approaches that identify 
change points as the segment boundaries that optimise a 
statistical cost function across the segmented time series. \textit{IGTS} \cite{sadri2017information} and \textit{OnlineIGTS} \cite{zameni2019unsupervised}
estimated change points by proposing top-down and dynamic programming approaches to search for the  
boundaries that maximised the information gain of the segmented time series. \textit{GGS} \cite{hallac2018greedy} proposed an online CPD approach that used a greedy search to identify the boundaries that maximised the regularised likelihood estimate of the segmented Gaussian model.

Another broad category of CPD methods exploit the temporal shape patterns of time series. 
\textit{FLOSS} was proposed to detect change points by identifying the positions within
the time series associated with a salient change in its shape patterns \cite{gharghabi2018domain}. Authors of \cite{xia2020robust} proposed a motif discovery approach in order to extract rare patterns that can distinguish separate segments \cite{huang2014detecting}. 
Recently, \textit{ESPRESSO} \cite{deldari2020Espresso} proposed a hybrid CPD approach that exploit both the temporal shape pattern and statistical distribution of time series. It was shown that the hybrid model was able to detect change points across a diverse range of time series datasets with greater accuracy than purely statistical or temporal shape based methods. 



Deep learning based CPD methods have also recently been proposed.
The authors of \cite{de2020change} used an AutoEncoder for CPD by exploiting peaks in the reconstruction error of the encoded representation. Kernel Learning Change Point Detection, \textit{KL-CPD} \cite{chang2019kernel}, is a state-of-the-art end-to-end CPD method which solves the problem of parameter tuning in kernel-based methods, by automatically learning the kernel parameters and combining multiple kernels to capture different types of change points. \textit{KL-CPD} utilised a two-sample test for measuring the difference between contiguous sub-sequences. \textit{KL-CPD} was shown to significantly outperform other deep learning and non deep learning CPD methods.


\begin{figure*}[]

    \centering
    \includegraphics[width=0.7\linewidth]{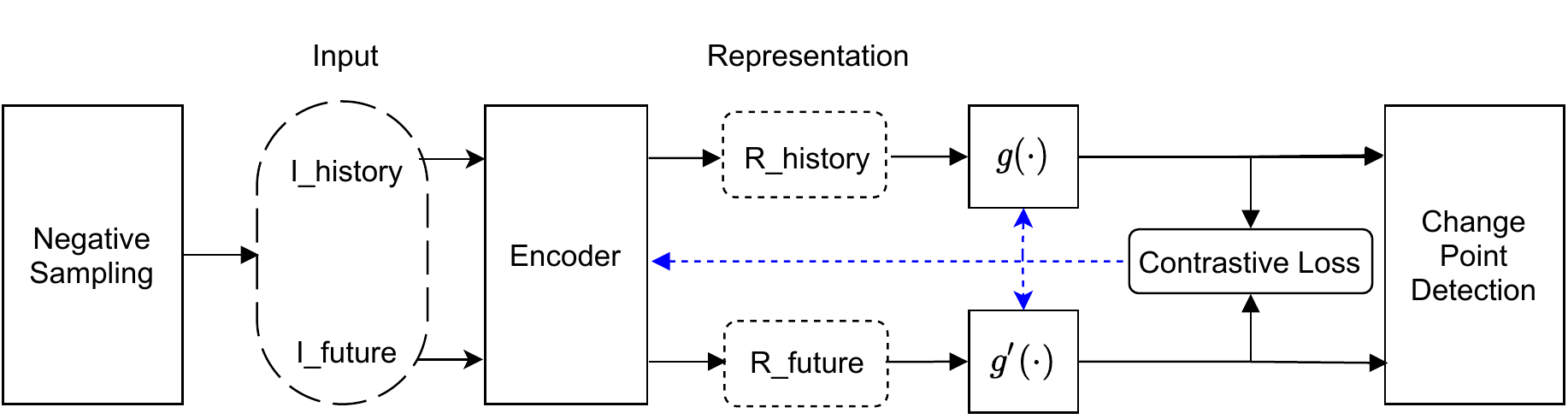}
    
    \caption{Illustration of the overall architecture of our~\TSCP. Blue dash arrows indicate the back propagation in the training phase.}
    \label{fig:overview}
\end{figure*}

CPD is also useful in video processing applications for summarising video, extracting segments of interest \cite{wei2018sequence}, and automatic caption generation and synchronisation \cite{aakur2019perceptual,ding2018weakly,farha2019ms}. Existing video segmentation approaches are commonly supervised and benefit from having knowledge of the order of actions. In contrast, \cite{aakur2019perceptual} proposed an auto-regressive model to predict the next video frames based on the most recently seen frames. Abrupt increases in the prediction error were then used to detect the segment boundaries.

\subsection{Representation Learning}
In recent years, self-supervised representation learning has been used to 
capture informative and compact representations of video \cite{aakur2019perceptual,mounir2020temporal}, image \cite{chen2020simclr,henaff2020data}, text\cite{oord2018representation}, and time series \cite{saeed2019multi,lu2020segmentation,saeed2020iot, franceschi2019unsupervised} data.

\subsubsection{Contrastive Learning}
Contrastive learning is an approach used to formulate what makes
the samples in a dataset similar or dissimilar using 
a set of training instances composed of
positive sample pairs (samples considered to be similar in some sense) 
and negative sample pairs (samples considered to be different). A representation
is learnt to bring the positive sample pairs closer together and to further
separate negative sample pairs within the embedding space.
Contrastive loss \cite{chopra2005contlosslearning} and Triplet loss \cite{weinberger2009distance} are the most commonly used loss functions. In general, the triplet loss function outperforms the contrastive loss function because it considers the relationship between positive and negative pairs, whereas the positive and negative pairs are considered separately in the contrastive loss function. Triplet loss, however, only considers one positive and one negative pair of instances at a time. Both functions suffer from slow convergence and require expensive data sampling methods to provide informative instance pairs, or triplets of instances, that accelerate training \cite{sohn2016improved}.
To solve the aforementioned problems, Multiple Negative Learning loss functions have been proposed to consider multiple negative sample pairs simultaneously. \textit{N-Paired loss} \cite{sohn2016improved} and \textit{infoNCE} based on \textit{Noise Contrastive Estimation} \cite{gutmann2010noise,mnih2013learning}  are examples of recent multiple negative learning loss functions. These approaches, however, require computationally expensive sampling approaches to select negative sample instances for training. This issue of complexity has been addressed by Hard Negative Instance Mining, which has been shown to play a critical role in ensuring contrastive cost functions are more efficient \cite{duan2019deep,wu2017sampling}. A number of sampling strategies have been proposed, including hard negative sampling \cite{simo2015discriminative}, semi-hard mining \cite{schroff2015facenet}, distance weighted sampling \cite{wu2017sampling}, hard negative class mining \cite{sohn2016improved}, and rank-based negative mining \cite{wang2019ranked}.

\subsubsection{Contrastive-based Representation Learning}

Most existing work on representation learning focus upon natural language processing \cite{oord2018representation} and computer vision \cite{chen2020simclr,henaff2020data} domains.
Howerver, to the best of our knowledge, it is the first time contrastive learning has been used for change point detection.

There is a few works that investigates 
the use of representation learning with multivariate time series. The authors of \cite{franceschi2019unsupervised} proposed a general-purpose approach to learn representations of variable length time series using a deep dilated convolutional network (WaveNet \cite{oord2016wavenet}) and an unsupervised triplet loss function based on negative sampling.

Contrastive predictive coding, \textit{CPC} \cite{oord2018representation}, uses auto-regressive models to learn representations within a latent embedding space. The aim of CPC is to learn within an abstract, global representation of the signal as opposed to a high dimension, lower level representation. The authors demonstrated it could learn effective representations of different data modalities such as images, text and speech for downstream modelling tasks. Firstly, a deep network encoder was used to map the signal into a lower dimension latent space before an auto-regressive model was then applied to predict future frames. A contrastive loss function maximised the mutual information between the density ratio of the current and future frames. \textit{CPCv2} \cite{henaff2020data} replaced the auto-regressive RNN of \textit{CPC} with a convolutional neural network (CNN) to improve the quality of the learnt representations for image classification tasks. 

\section{Method}
\subsection{Problem Definition}
\label{sec:problem}
Given a multivariate time series $\{X_{1}, X_{2},...,X_{T}\}$ of $T$ observations, where the vector $X_{i} \in \R^{d}$, we attempt to estimate the times ($t$) that are associated with a change in the time series properties. We define change points (or segment boundaries) as the time points in future can not be anticipated from the data before this point. Hence, the dissimilarity between future representation and anticipated representations can be used as a measure to detect transition to the next segment.



\begin{figure*}[]
    \centering
    \includegraphics[width=0.9\linewidth]{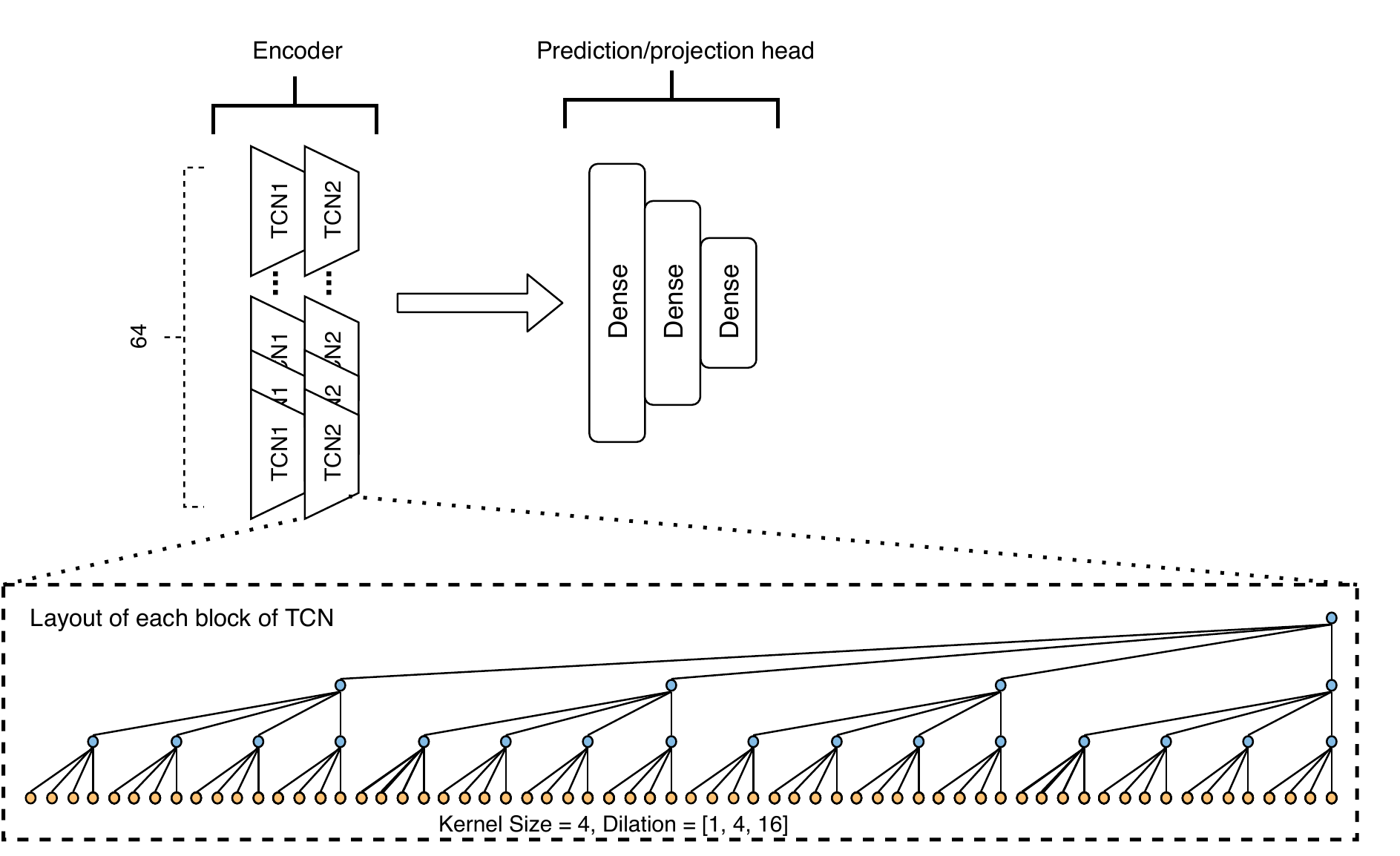}
    \caption{The encoder architecture for \TSCP\. We use two stacks of TCN with kernel size 4 and dilation sizes of 1, 4, and 16 followed by three Dense layers as the prediction head. }
    \label{fig:TCN_model}
\end{figure*}

\subsection{\TSCP\  Overview}

There are change point and temporal anomaly detection methods for video \cite{aakur2019perceptual} and time series \cite{kokkonen2019network} that use an auto-regressive model for prediction. In these approaches, change points are detected at samples associated with a salient increase in the prediction error. However, since the prediction error is highly dependent upon the distribution of the data, we propose to use representation learning to extract a compact latent representation that is invariant to the original distribution of the data. Figure \ref{fig:main_idea} illustrates the main idea behind our \TSCP\ approach. We hypothesize that this approach is much more effective to detect change points because the embedding space that is extracted from contiguous time intervals are likely to be dependent upon the same shared information.

Here we adopt a similar approach to the CPC \cite{oord2018representation,henaff2020data} method to learn a representation
that maximises the mutual information between consecutive time windows. Firstly, an auto-regressive deep convolution network, WaveNet \cite{oord2016wavenet}, was employed to encode each of the time series windows. Secondly, a 3-layer fully connected network was employed on top of this encoding to produce a more compact, embedded representation. The cosine similarity was computed between the embeddings of consecutive time windows in order to estimate the change points. The time intervals associated with smaller similarity values had a higher likelihood of being change points.
A contrastive learning approach was used to train the encoder by using a single pair of contiguous time windows (positive pair) and a set of window pairs that were separated across time (negative pairs) within each batch. 

We applied the $N-paired loss$ metric \cite{sohn2016improved} (which is described in section \ref{subsec:costfn}) to maximise the mutual information between 
the positive pairs amongst the set of negative pairs of samples.  



Figure \ref{fig:overview} shows the overall architecture of the proposed method. In the following section we will describe the main modules in the following order: 1) Representation learning, 2) Negative sampling, and 3) Change point detection.
\par

\subsection{Representation Learning}
At the core of the \TSCP\ approach is an encoder that maps 
pairs of contiguous time windows 
into a compact embedding representation. This representation
was trained to learn about the concept of similarity 
over short temporal scales by maximising the 
mutual information between the pairs of adjacent
time windows.  
We employ the auto-regressive deep convolution network, WaveNet \cite{oord2016wavenet}, to learn our encoded representation. We do not use an LSTM to encode the time series, given it has been shown that temporal convolutional networks (TCN) can often produce superior prediction performance with sequential data \cite{henaff2020data} and are generally easier to train. 


Figure \ref{fig:TCN_model} illustrates the encoder architectures. It consists of two blocks of TCN with 64 kernel filters of size 4 and three layers of dilation with respective rates of 1, 4 and 16. The TCN is then followed by a simple three-layer projection head with ReLU activation function and batch normalisation. The modified illustration of TCN layer\footnote{We acknowledge the main illustration and TCN implementation: https://github.com/philipperemy/keras-tcn} is shown in the figure. 

Pairs of history and future time windows are fed into an encoder. A projection head is used (shown as $g(\cdot)$ and $g^{\prime}(\cdot)$ in Figure \ref{fig:overview}, respectively) to map each
window encoding into a lower dimension space. To this end, an MLP neural network with three hidden layers was used. 

Contrastive learning used pairs of 
history and future windows for training an embedded representation. Two different types of time window pairs were contrasted for training. Each training instance was comprised of a
positive sample pair of contiguous time intervals and
a set of negative sample pairs with intervals separated across time. In the next subsection, we define the $InfoNCE$ cost function that will be used for representation learning.

\subsubsection{Cost function}
\label{subsec:costfn}
 We applied the $InfoNCE$ loss function that is based upon Noise Contrastive Estimation \cite{mnih2013learning}, which was originally proposed for natural language representations but has also recently been adopted for image representation learning techniques \cite{henaff2020data, oord2018representation, chen2020simclr}. 
 
 The $InfoNCE$ cost function is defined to maximise the mutual information between consecutive time windows. A single positive pair of time adjacent intervals $(h_i,f_i)$, the history window (${h_i}$) and future window (${f_i}$), and a set of $K-1$ negative pairs ($(h_i,f_j)_{j\ne i}$) where the intervals ${h_i}$ and ${f_j}$ were well separated in time across the sequence.  


Using the $InfoNCE$ loss function, we calculate the probability $\rho_{i}$ of the positive sample pair in each batch using the scaled-$Softmax$ function:

\begin{equation}
 \label{eq:prob}
  \rho_{i} = {\frac{exp(Sim(h_i,f_i)/\tau)}{\sum_{j=1}^{K}{exp(Sim(h_i,f_j)/\tau)}}}\ %
  \end{equation}

where ${\tau}$ is a scaling parameter and $Sim$ is the cosine similarity between each pair of data embeddings. The final loss is calculated with the binary cross-entropy function over the probabilities of all $K$ positive pairs belonging to the training batch. Since the probabilities of the positive sample pairs are computed using the similarity scores of the negative sample pairs in (\ref{eq:prob}), the cross entropy loss function can be simplified to:

\begin{equation}
    \label{eq:bin_loss} 
    \mathbf{\mathscr{L}} = - \sum_{i,j}{y_{ij} log(\rho_{i}) + (1-y_{ij}) log(1-\rho_{i})}
\end{equation}

\begin{equation}
\centering
    \label{lbl}
    \mathscr{y}_{ij} = \begin{cases}
  1 & \text{if $i=j$} \\
  0 & \text{if $i \neq j$}\\
\end{cases}
\end{equation}

\begin{equation}
    \label{loss}
    \mathbf{\mathscr{L}} = \sum_{i}{-log(\rho_{i})}
\end{equation}


\subsubsection{Negative Sampling}
\begin{figure}[!htbp]
    \centering
    \includegraphics[width=\linewidth]{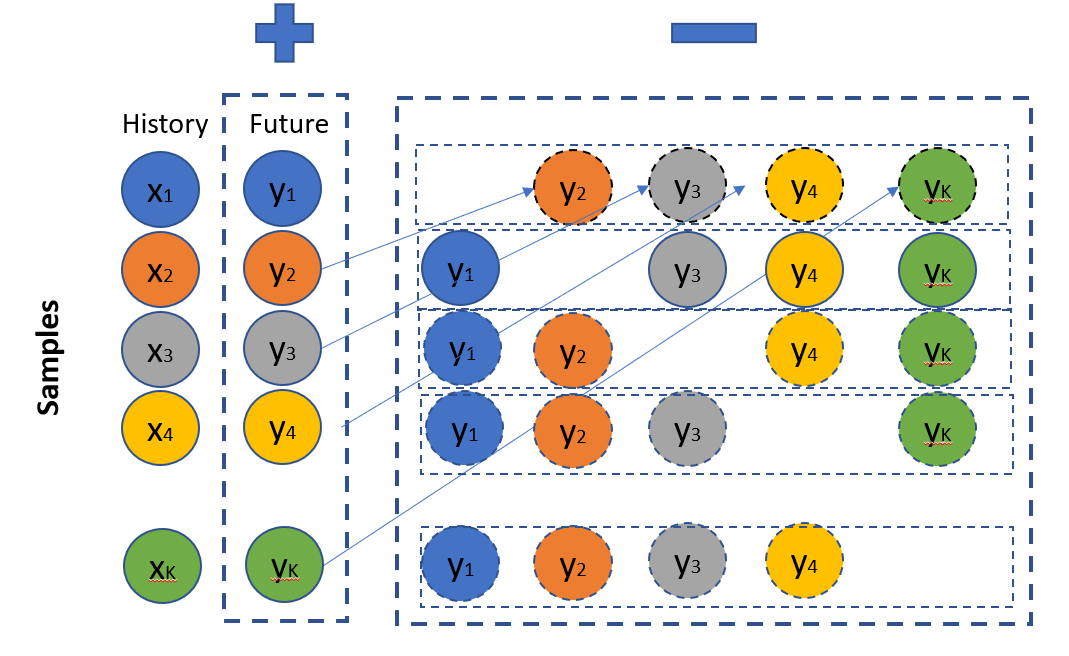}
    \caption{Batch construction}
    \label{fig:sampling}
\end{figure}

Following on from the hard negative class mining approach in \cite{sohn2016improved}, we propose a simpler sampling strategy where positive sample pairs are randomly sampled and used to construct the negative sample pairs for each batch. 
Figure \ref{fig:sampling} depicts the process of batch construction in our model. We choose $K$ random pairs of contiguous windows $(h_i,f_i)$ as the positive pairs in each training batch. Each pair must adhere to the constraint of being a minimum temporal distance from the other pairs. This minimum temporal distance constraint is used to enable each batch to adopt the future windows of the other $K-1$ positive pairs as negative pairs, given they are guaranteed to be sufficiently separated from the history window in the batch's own positive pair. The intuition is that time series are commonly non-stationary, and hence, windows that are temporally separate from one another are likely to exhibit far weaker statistical dependencies than adjacent windows. We need to set the threshold of minimum temporal distance based upon the time series application being considered.

\begin{figure*}
\centering
\begin{minipage}[t]{.48\textwidth}
        
        \subfigure{\includegraphics[width=\linewidth]{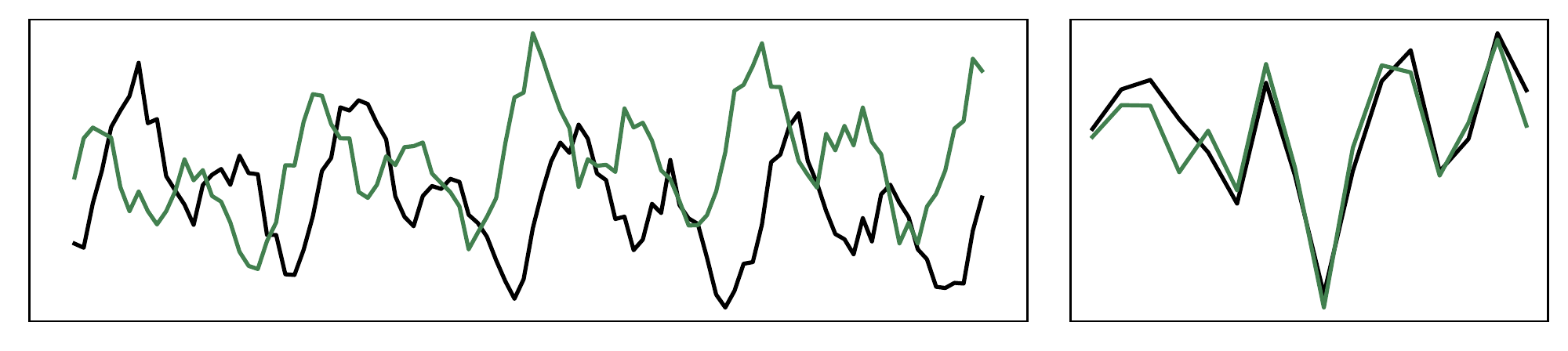}}
        \subfigure{\includegraphics[width=\linewidth]{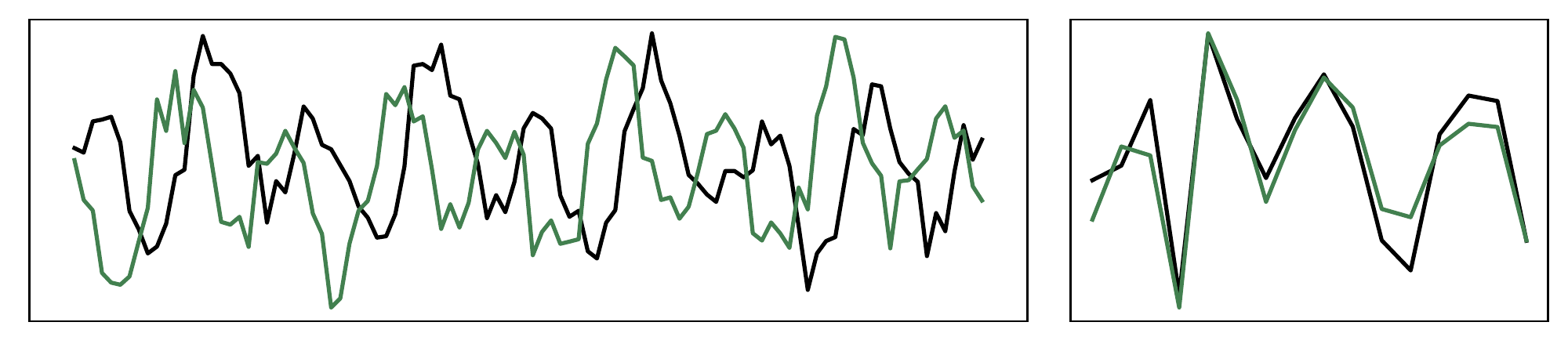}}
        \subfigure{\includegraphics[width=\linewidth]{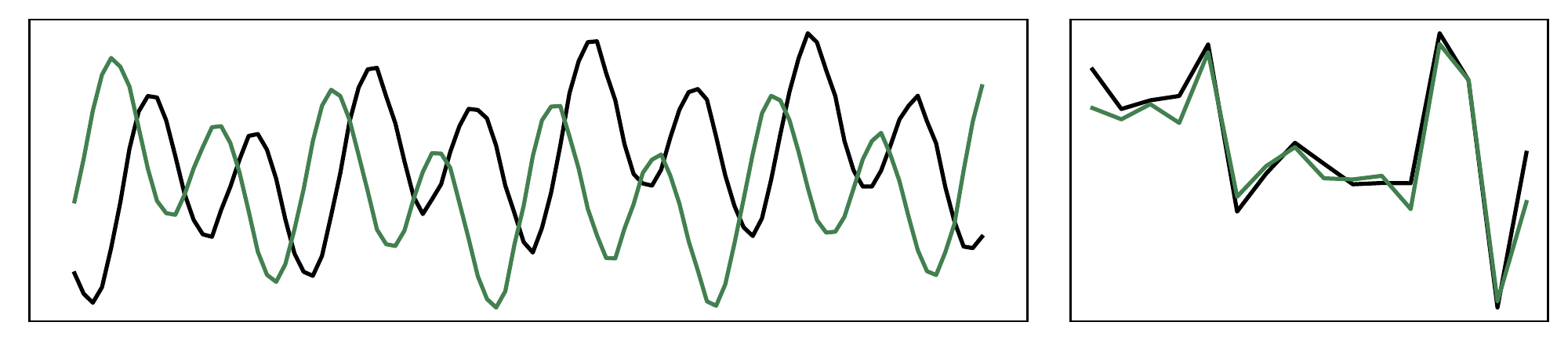}}
        \caption{Three examples of positive pairs of samples (left, length = 100) and their corresponding embedding (right, length = 16). The positive pairs are subsequent intervals of the time series.}\label{fig:positive}
\end{minipage}\qquad
\begin{minipage}[t]{.48\textwidth}
        \subfigure{\includegraphics[width=\linewidth]{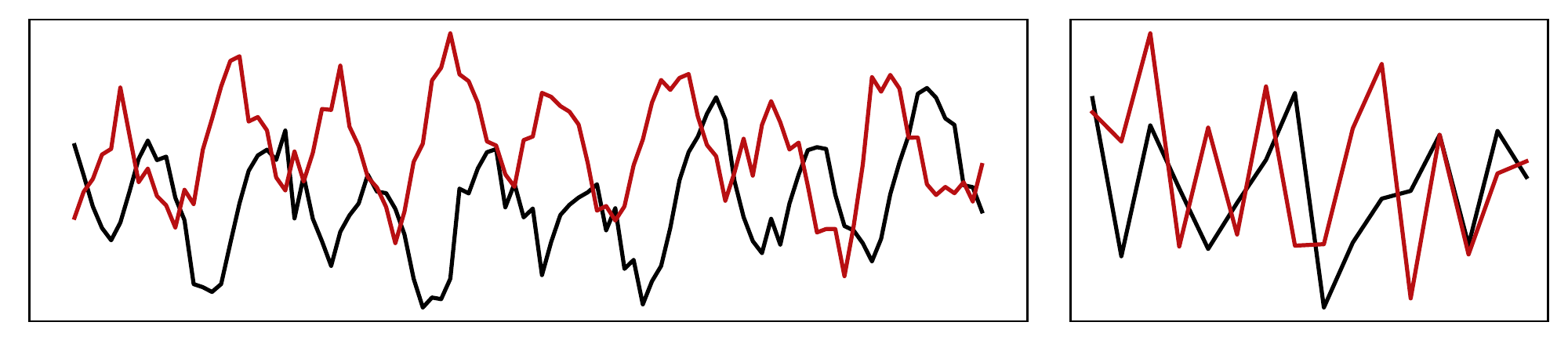}}
        \subfigure{\includegraphics[width=\linewidth]{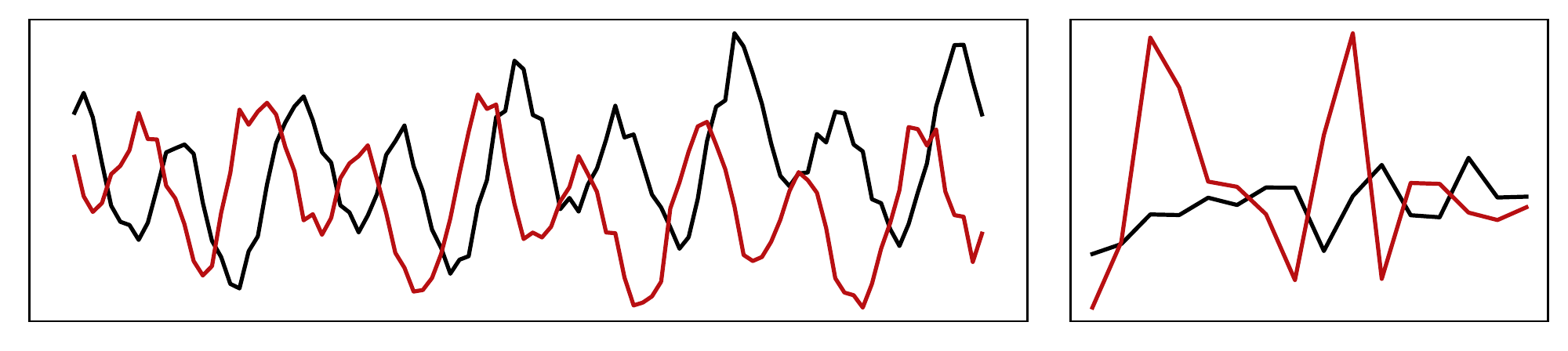}}
        \subfigure{\includegraphics[width=\linewidth]{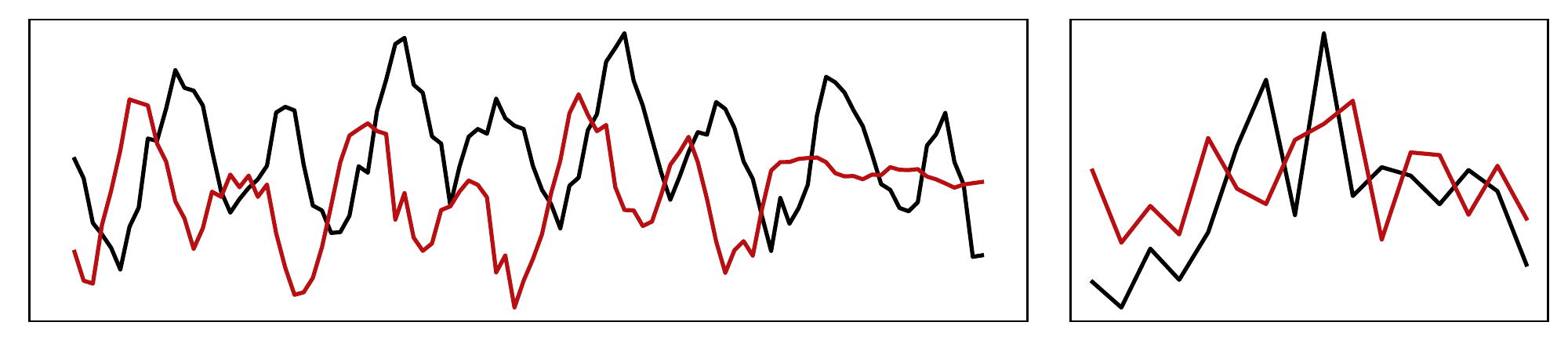}}
        \caption{Three examples of negative pairs of samples (left, length = 100) and their corresponding embedding (right, length = 16). The negative pairs correspond to time intervals of the time series that well separated in time. A change point or anomaly has occurred somewhere in either history or future frame of the negative pairs.}\label{fig:negative}
\end{minipage}
\end{figure*}

Consequently, we can select positive and negative sample pairs with a relatively low complexity relative to the other negative mining approaches mentioned in Section \ref{sec:related}. Figure \ref{fig:positive} and \ref{fig:negative} show examples of time windows belonging to the positive and negative pairs, respectively, and their corresponding embedding vectors. 

\subsection{Change Point Detection Module}
We hypothesize that when a change point intersects a pair of history and future windows, their associated 
embeddings will be distributed differently. Consequently, in order to detect change points from
the time series being tested, we transform 
pairs of history and 
future windows into a compact embedding and 
compute the cosine similarity ($Sim(h_i,f_i)$) between
the embedding pairs across the time series being tested. The difference between the 
cosine similarity and moving average of the cosine similarity
was computed and a peak finding algorithm was applied to find 
local maxima in the difference function (increase in difference function is associated with decrease in similarity metric). The time intervals
associated with these local maxima are considered as the change point estimates. 

\begin{figure}[!htbp]
    \centering
    \includegraphics[width=\linewidth]{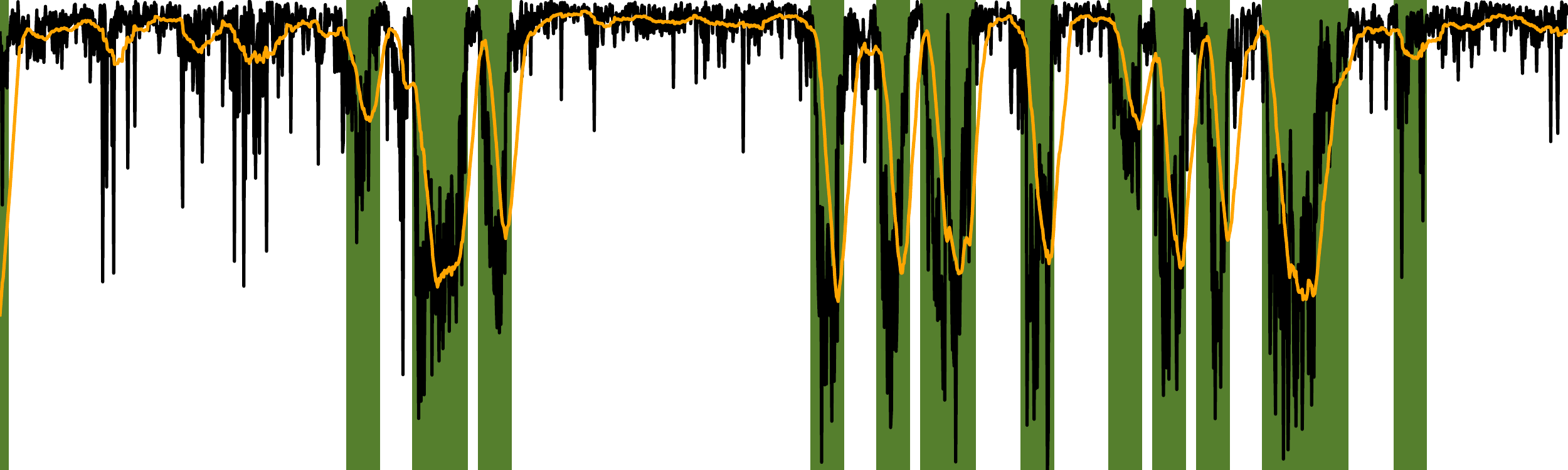}
    \caption{An example of detecting change points based upon the cosine similarity between consecutive 
    window pairs. The black line shows the cosine similarity between subsequent time intervals in Benchmark-4 of the Yahoo! dataset \cite{yahoodataset}. The green areas highlight the intervals with change points and the yellow line shows the moving average of the cosine similarity for the previous $W$ intervals.}
    \label{fig:cpd_example}
\end{figure}

Figure \ref{fig:cpd_example} shows an example of the cosine similarity between the latent embeddings of the history and future windows within a time series. The green areas show the interval pairs $(h_i,f_i)$ which contain a change point for a subset of the Benchmark-4 of the Yahoo! dataset \cite{yahoodataset}. It is clear that the local minima of the difference between the cosine similarity of each interval pair $(h_i,f_i)$ and the average similarity across recent intervals pairs coincide with true change points. 

\section{Experiments}
\label{sec:experiments}
In this section, we present our evaluation of the proposed \TSCP\ method. Firstly, we introduce the datasets and  outline the baseline CPD methods used in our experiments. A sensitivity analysis of the \TSCP\ method is presented along with a performance comparison with baseline CPD methods. \TSCP\ is implemented using Tensorflow 2.2.0 and python 3.7.2.

\subsection{Datasets}
We show the effectiveness of our method across a diverse range of applications that include web service traffic analysis, human activity recognition and mobile application usage analysis.  

\begin{itemize}
\item \textbf{Yahoo!Benchmark}\footnote{Yahoo Research Webscope dataset, S5 - A Labeled Anomaly Detection Dataset, version 1.0,https://webscope.sandbox.yahoo.com/} \cite{yahoodataset}. The Yahoo! benchmark dataset is one of the most widely cited benchmarks for anomaly detection. It contains time series with varying trend, seasonality, and noise including random anomaly change points. We used all 100 time series of the fourth benchmark, as it is the only portion of the dataset that includes change points. 

\item \textbf{HASC} \footnote{http://hasc.jp/hc2011} \cite{nobuo2011hasc,kawaguchi2011hasccorpus}. The HASC challenge 2011 dataset provides human activity data collected by multiple sensors including an accelerometer and gyrometer. We used a subset of the HASC dataset (The same subset used by recent state-of-the-art method, \textit{KL-CPD}) including only 3-axis accelerometer recordings. The aim of detecting change point detection with this dataset is to find transitions between physical activities such as \textit{"stay", "walk", "jog", "skip", "stair up", "stair down"}.

\item \textbf{USC-HAD} \footnote{http://sipi.usc.edu/had} \cite{zhang2012usc}. USC-HAD dataset includes twelve human activities that were recorded separately across multiple subjects. Each human subject was fitted with a 3-axis accelerometer and 3-axis gyrometer that were attached to the front of the right hip and sampled at 100Hz. Activities were repeated five times for each subject and consisted of: \textit{"walking forward", "walking left", "walking right", "walking upstairs", "walking downstairs", "running forward", "jumping up", "sitting", "standing", "sleeping", "elevator up"}, and \text{"elevator down"}. We randomly chose 30 activities from the first six participants and stitched the selected recordings together in a random manner. In the experiments undertaken in this paper, only the data from the accelerometer was used.\par

\end{itemize}
Table \ref{tab:dataset_info} outlines the properties of each dataset.

\begin{table}[htb]
    \centering
    \caption{The properties of the three datasets used in our experiments. 
    T is
    the total number of samples, \#sequences is the quantity of 
    time series, \#channels represents the 
    time series dimensionality 
    and \#CP is the total number of change points in each dataset. }
    \begin{tabular}{c|cccc} 
\hline
dataset          & T      & \#sequences & \#channels & \#CP  \\ 
\hline
Yahoo! Benchmark & 164K & 100         & 1          & 208   \\
HASC             & 39K  & 1           & 3          & 65    \\
USC-HAD          & 97K  & 6           & 3          & 30    \\
\hline
\end{tabular}
    
    \label{tab:dataset_info}
\end{table}

\begin{figure*}[]
\centering
        \subfigure{\includegraphics[width=.35\linewidth]{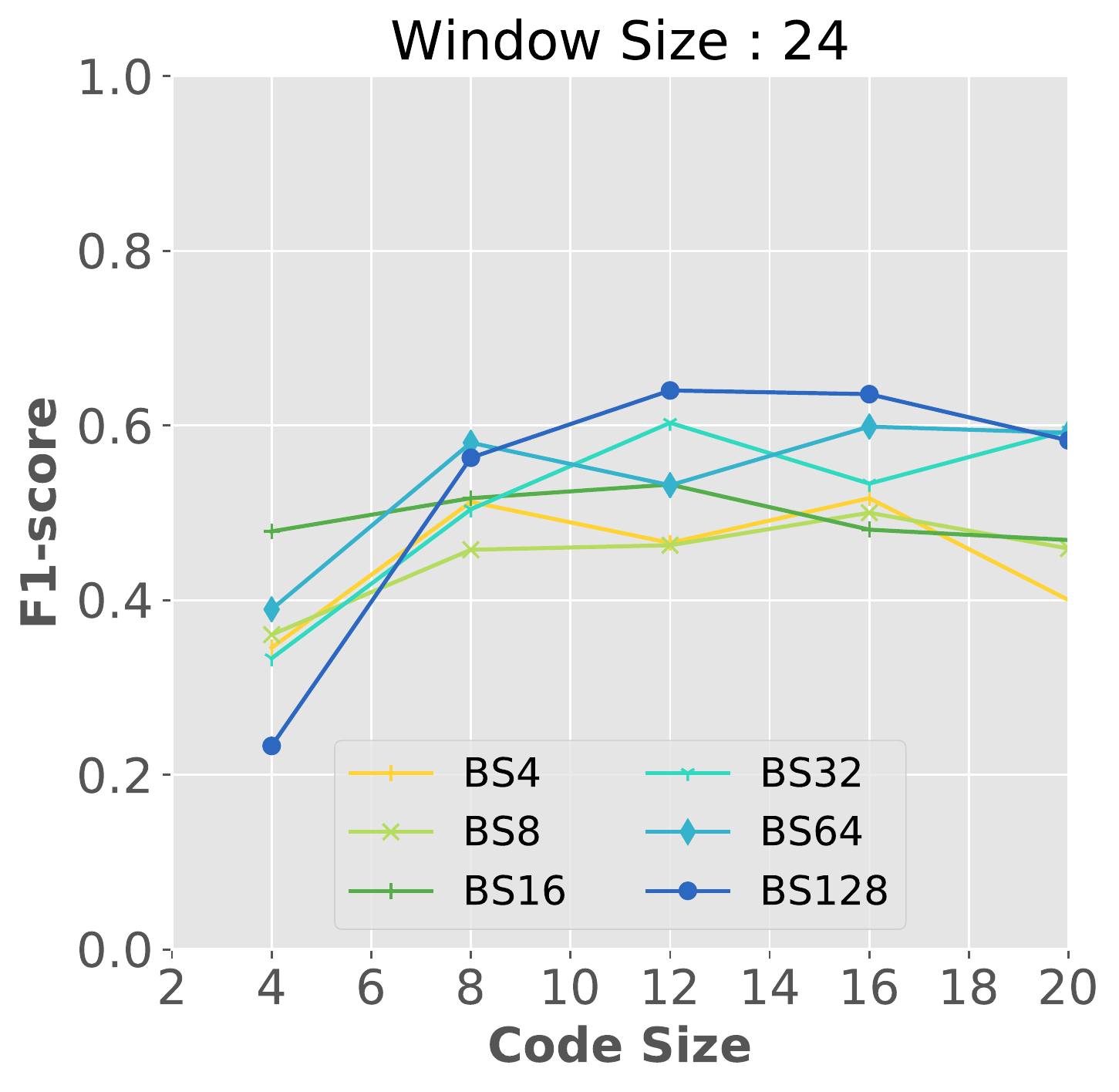}}
        \subfigure{\includegraphics[width=.35\linewidth]{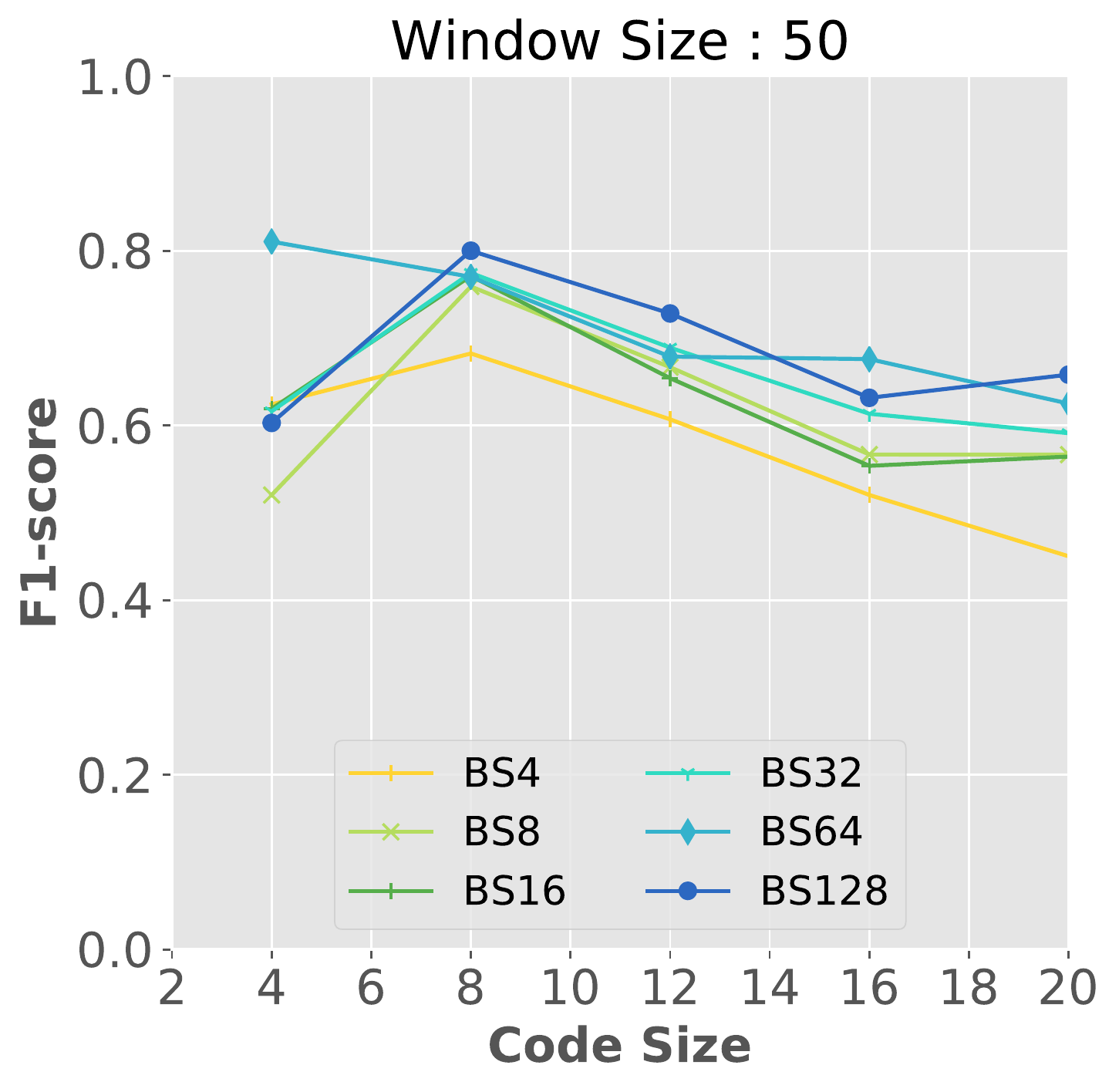}\label{fig:win50}}
        \hfill
        \subfigure{\includegraphics[width=.35\linewidth]{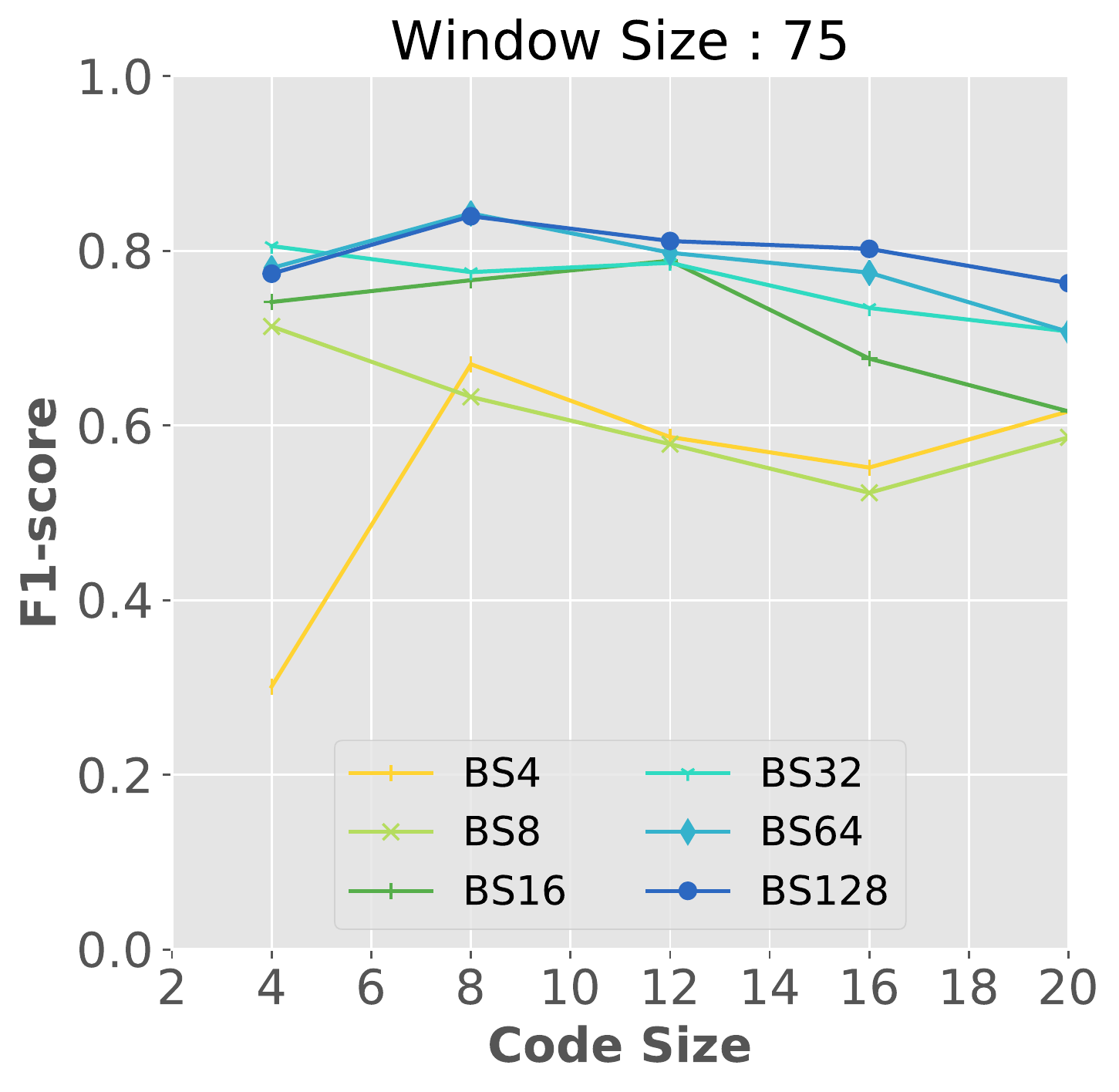}\label{fig:win75}} 
        \subfigure{\includegraphics[width=.35\linewidth]{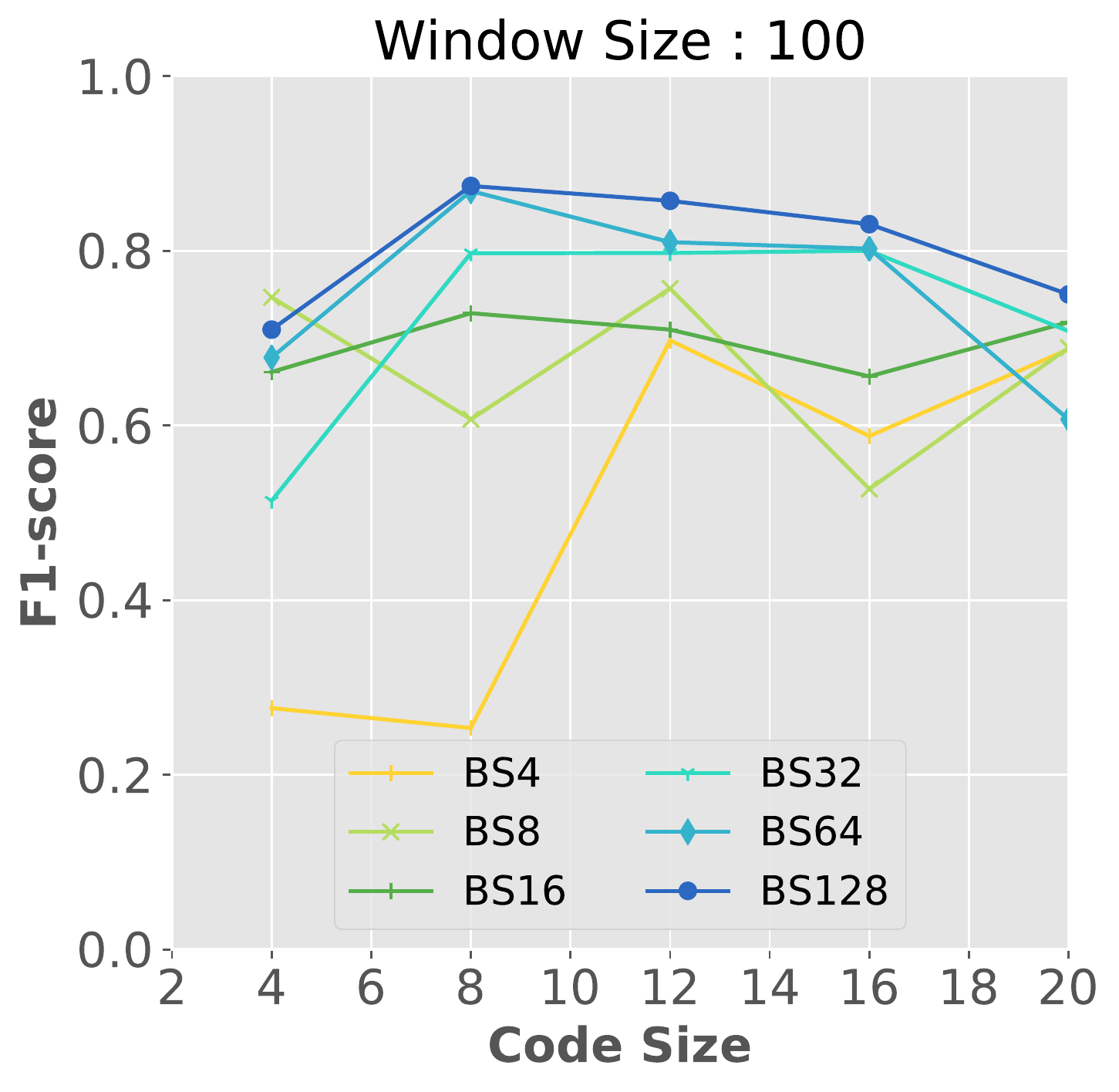}}\\
        
        \caption{A sensitivity analysis of the  the Yahoo!Benchmark dataset with respect to the code size, batch size and four window sizes of: (a) 24 samples  (b) 50 samples (c) 75 samples and (d) 100 samples.  }
        \label{fig:sensitivity}
\end{figure*}

\subsection{Baseline Methods}
The performance of the proposed \TSCP\ method was compared against five state-of-the-art unsupervised change point detection techniques that included \textit{ESPRESSO} \cite{deldari2020Espresso}, \textit{FLOSS} \cite{gharghabi2018domain}, \textit{aHSIC} \cite{Yamadaahsic}, \textit{RuLSIF} \cite{liu2013change}, and \textit{KL-CPD} \cite{chang2019kernel}. 
To avoid inconsistencies and implementation errors, and to provide a fair comparison, baseline methods were evaluated using the publicly available source code. 


Kernel specific parameters were used by the \textit{RuLSIF} and \textit{aHSIC} methods. For \textit{RuLSIF}, the regularisation constant was set to a value of 0.01, as suggested in \cite{liu2013change}. For \textit{aHSIC}, the regularisation constant and the kernel bandwidth parameters were set to values of 0.01 and 1, respectively, as specified in \cite{Yamadaahsic}.

Detection performance was compared across a range of window sizes that were unique to each dataset based on its sampling rate. 
As a deep learning based method, \textit{KL-CPD} required several hyper-parameters to be tuned; window size, batch size and learning rate. To enable a fair comparison with the other methods, a grid search was performed across the sets of hyper-parameter values. Only the hyper-parameter configuration that provided the highest rate of true positives was presented. We used the same evaluation approach as undertaken with \textit{KL-CPD} \footnote{KL-CPD source code: https://github.com/OctoberChang/KL-CPD\_code} to calculate the F1-score. The remaining parameters were set according to the values specified in \cite{chang2019kernel}. Although the training process of \textit{KL-CPD} was unsupervised, the method still required ground truth labels to be used to fine-tune the 
model hyper-parameters during the validation phase. For the \textit{FLOSS} and \textit{ESPRESSO} methods, we used the z-normalised euclidean distance as the similarity metric, as suggested by authors of their underlying structure they used in \cite{zhu2016matrix}.

\subsection{Evaluation Metrics}
The performance of models were evaluated with respect to the F1-score. 
The error margin in which a change point can be detected is an important factor in evaluating the performance of each CPD method \cite{Aminikhanghahi2017}. Hence, we report the F1-scores of each dataset for
the three different detection margins specified in Table \ref{tab:compare_baseline}. 

Each change point estimate was defined as a true positive when it was located within the specified error margin of the ground truth change point. When multiple change point estimates were located within the error margin of the the ground truth change point, only the closest estimate was considered to be a true positive. The remaining estimates were considered to be false positives. Ground truth change points without any estimates that fell within the specified error margin were considered to be false negatives.

\subsection{Fine-Tuning and Sensitivity Analysis}

In this section, we have done extensive experiments to analyse the sensitivity of our proposed method to:
\begin{itemize}
    \item \textbf{Window size} is the length of the history and future intervals. We consider a range of values that are dependent upon the particular application and sampling rate of the dataset. The window size should be large enough to encapsulate the properties of the time series but not too large to encompass multiple change points. We select window sizes of 1,2,3 and 4 days for the Yahoo!Benchmark dataset and 0.5,1,2 and 4 seconds for the HAR dataset.
    \item \textbf{Batch Size} specifies the number of training instances ($K$) that were processed before the model was updated. It also specifies the number of negative pairs ($K-1$) used in each training instance. The batch size was selected to range between 4 and 128 samples. 
    \item \textbf{Code Size} specifies the length of the embedding vector that is extracted from the encoder network. The range of code sizes that were selected ranged between 4 to 20 dimensions.
\end{itemize}

Window size is the only input parameter to investigate for the ESPRESSO and FLOSS methods and the 
main parameter for the \textit{RulSIF} and \textit{aHSIC} methods. For the deep learning based methods, including the proposed method, we also investigate the performance of a number of parameters including the window size, batch size, code size and learning rate. 
Figure \ref{fig:sensitivity} compares the \TSCP\ performance (with respect to the F1-score) for the Yahoo!Benchmark dataset across the different parameter settings. 

\subsubsection{Window size} 
Since the Yahoo!Benchmark dataset is sampled hourly and the minimum length between two consecutive change points is approximately 160 samples, we consider a range of window sizes between 24 (1 day) to 100 (~4 days). Figure  \ref{fig:sensitivity} shows there was a monotonic increasing relationship between the window size and detection performance averaged across the code size and batch size. It was hypothesized that longer windows possess the highest F1 scores, given they encapsulate additional properties of the time series into modelling.

\subsubsection{Batch Size}
We varied the batch size with respect to the set of $\{4, 8, 16, 32, 64, 128\}$ dimensions to investigate its impact upon detection performance. 
Figure \ref{fig:sensitivity} shows there was a monotonic increasing relation between the 
batch size and detection performance when averaged across the code size 
and window size. There are particular situations for the smallest code size, however, where the largest batch sizes had inferior detection performance to the smaller batch sizes. We hypothesize such situations can occur given larger batches are more likely to generate 
false negative samples from the time series datasets and smaller code size are too short to represent all informative features of data. False pairs of negative samples are pairs of time windows that are considered to be false instances, but are found to be similarly distributed. Whilst the negative sample pairs are constrained to be intervals that are temporally separate from one another, time series are often comprised of patterns (and their associated semantic classes) that repeat at different, non-contiguous positions within the sequence. Consequently, using contrastive learning to separate these false pairs of negative samples within the embedding space can degrade detection performance.

\subsubsection{Code Size}
In contrast to many representation learning approaches, we investigated how the embedding dimensionality affected the detection performance. We varied the code size from 4 to 20 dimensions which equates to representing between 4\% to 83\% of a window from the experiment. As shown in Figure \ref{fig:sensitivity}, the optimal code size was dependent upon the window size and batch size. In general, the smallest code size of 4 showed a relatively weak performance for each of the window sizes, given there was insufficient capacity to represent the key 
features of the time series to learn an effective representation. The relationship between detection performance and code size was not monontonic increasing, however, given the largest code sizes were often shown to be inferior to the more compact embeddings with a code size of between 8 and 12 dimensions. 

\begin{table*}
\caption{The performance of the proposed \TSCP\ method was compared to the other baselines methods across the Yahoo!Benchmark, HASC, and USC datasets. The \textbf{bold} and \underline{underlined} texts represent the methods with the first and second highest F1-scores, respectively. The detection margin is the maximum number of samples that an estimated change point can be from a ground truth change point to still be considered a True Positive. We present the highest F1-score of each method (for the best window size) and the F1-score of the methods averaged across all window sizes. }
\begin{tabular}{c|lcc|cc|cc}
\noalign{\hrule height 3pt}%
\hline

\multicolumn{1}{l|}{\multirow{2}{*}{Dataset}} & \multicolumn{1}{c}{Detection margin}  & \multicolumn{2}{c|}{24}  & \multicolumn{2}{c|}{50}  & \multicolumn{2}{c|}{75}  \\ \cline{2-8} 
\multicolumn{1}{l|}{}                         & \multicolumn{1}{l|}{Methods}       & Best Wnd      & F1-score & Best Wnd      & F1-score & Best Wnd      & F1-score \\ \hline
\multirow{6}{*}{Yahoo}                       
                                              & \multicolumn{1}{l|}{FLOSS}         & 45            & 0.2083   & 50            & 0.3375   & 55            & 0.4233   \\
                                              & \multicolumn{1}{l|}{aHSIC}         & 40            & 0.4092   & 40            & 0.4175   & 40            & 0.4392   \\
                                              & \multicolumn{1}{l|}{RuLSIF}        & 20            & 0.3175   & 20            & 0.3317   & 20            & 0.3700     \\
                                              & \multicolumn{1}{l|}{ESPRESSO}      & 50            & 0.2242   & 50            & 0.3400     & 70            & 0.4442   \\
                                              & \multicolumn{1}{l|}{KL-CPD}        & 24            & \underline{0.5787}   & 50            & \underline{0.5760}   & 75            & \underline{0.5441}   \\ \cline{2-8} 
                                              & \multicolumn{1}{l|}{\TSCP}        & 24            & \textbf{0.64}     & 50            & \textbf{0.8104}   & 75            & \textbf{0.8428}   \\ \hline
\noalign{\hrule height 3pt}%
\multirow{7}{*}{USC}                          & \multicolumn{1}{l|}{Detection margin} & \multicolumn{2}{c|}{100} & \multicolumn{2}{c|}{200} & \multicolumn{2}{c|}{400} \\ \cline{2-8} 
                                              & \multicolumn{1}{l|}{FLOSS}         & 100           & 0.2666  & 100           & 0.3666  & 400           & 0.4333      \\
                                              & \multicolumn{1}{l|}{aHSIC}         & 50         & 0.3333  & 50            & 0.3333  & 50            & 0.3999          \\
                                              & \multicolumn{1}{l|}{RuLSIF}        & 400           & 0.4666  & 400           & 0.4666  & 400           & 0.5333      \\
                                              & \multicolumn{1}{l|}{ESPRESSO}      & 100           & 0.6333  & 100           & \underline{0.8333} & 100 & \textbf{0.8333}  \\
                                              & \multicolumn{1}{l|}{KL-CPD}         & win:100, bs:4 & \underline{0.7426}   & win:200,bs:32 & 0.7180   & win:400,bs:16 & 0.6321   \\ \cline{2-8} 
                                              & \multicolumn{1}{l|}{\TSCP}     & win:100, bs:8 & \textbf{0.8235}  &  win:200, bs:8 & \textbf{0.8571} & win:400, bs:32  & \textbf{0.8333}         \\ \hline
\noalign{\hrule height 3pt}%
\multirow{7}{*}{HASC}                         & \multicolumn{1}{l|}{Detection margin} & \multicolumn{2}{c|}{60}  & \multicolumn{2}{c|}{100} & \multicolumn{2}{c|}{200} \\ \cline{2-8} 
                                              & \multicolumn{1}{l|}{FLOSS}         & 60            & 0.3088   & 60            & 0.3913   & 100           & 0.5430   \\
                                              & \multicolumn{1}{l|}{aHSIC}         & 40            & 0.2308   & 40            & 0.3134   & 40            & 0.4167   \\
                                              & \multicolumn{1}{l|}{RuLSIF}        & 200           & 0.3433   & 200           & 0.4999      & 200           & 0.4999      \\
                                              & \multicolumn{1}{l|}{ESPRESSO}      & 100           & 0.2879   & 60            & 0.4233   & 100           & \textbf{0.6933}   \\
                                              & \multicolumn{1}{l|}{KL-CPD}        & win:60,bs:4   & \textbf{0.4785}   & win:100,bs:4  & \textbf{0.4726}   & win:200,bs:64 & 0.4669   \\ \cline{2-8} 
                                              & \multicolumn{1}{l|}{\TSCP}        & win:60,bs:64  & \underline{0.40}   & win:100,bs:64 & \underline{0.4375}   & win:200,bs:64 & \underline{0.6316}   \\ \hline
\noalign{\hrule height 3pt}%

\end{tabular}

\label{tab:compare_baseline}
\end{table*}

\subsection{Baseline Comparison}
The performance of the proposed \TSCP\ method was compared to the five baseline methods across the 
three datasets. To enable a fair comparison between the methods, we performed a grid search
of the set of parameters associated with each method. For each method, the model 
with the best F1-score
and its corresponding parameters were presented in Table \ref{tab:compare_baseline}. 

\begin{figure}
    \centering
    \subfigure{\includegraphics[width=0.45\linewidth]{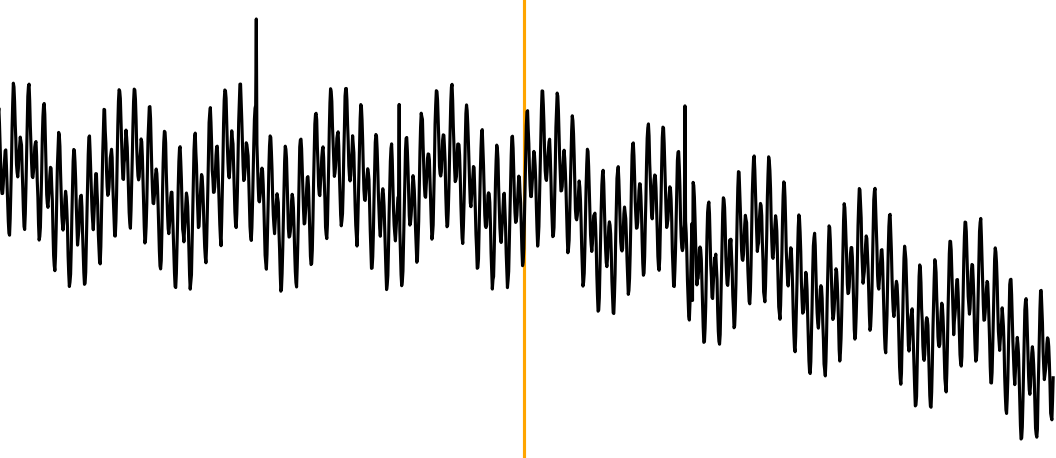}}
        \subfigure{\includegraphics[width=0.45\linewidth]{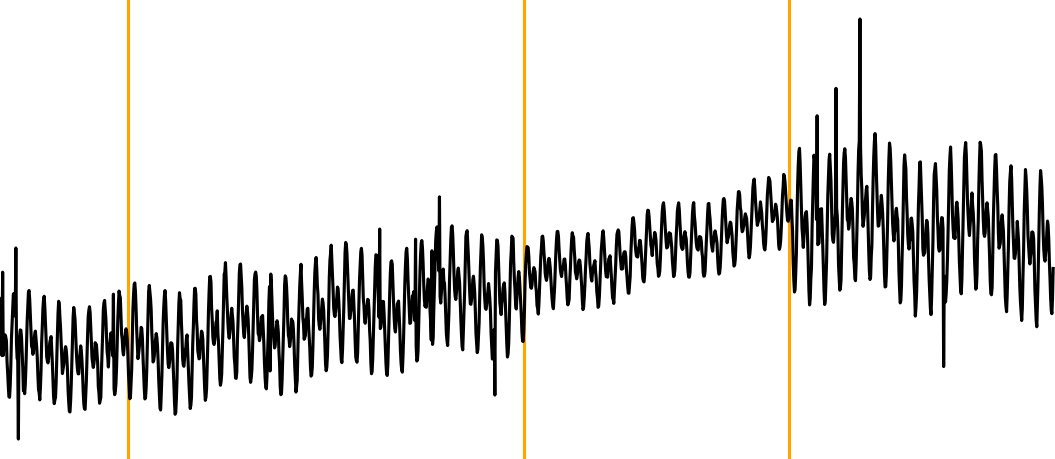}}
        \subfigure{\includegraphics[width=0.48\linewidth]{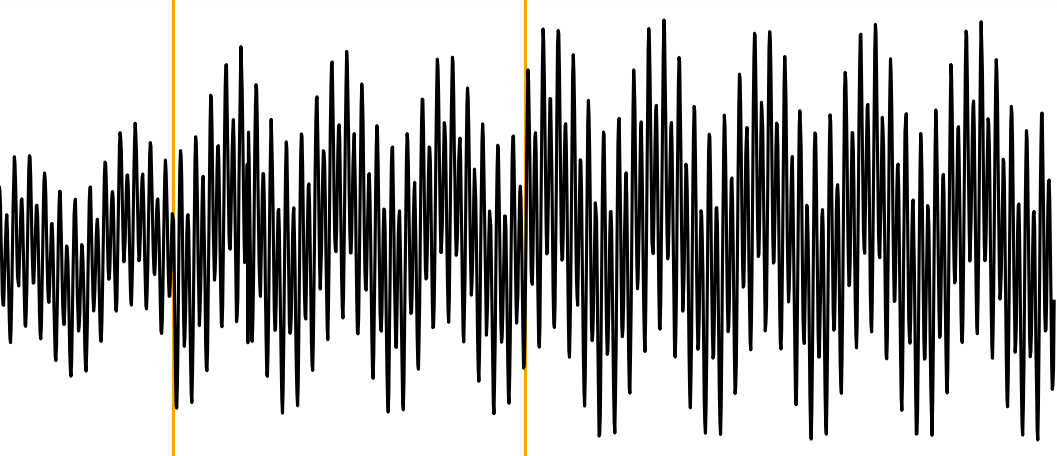}}
       \subfigure{\includegraphics[width=0.48\linewidth]{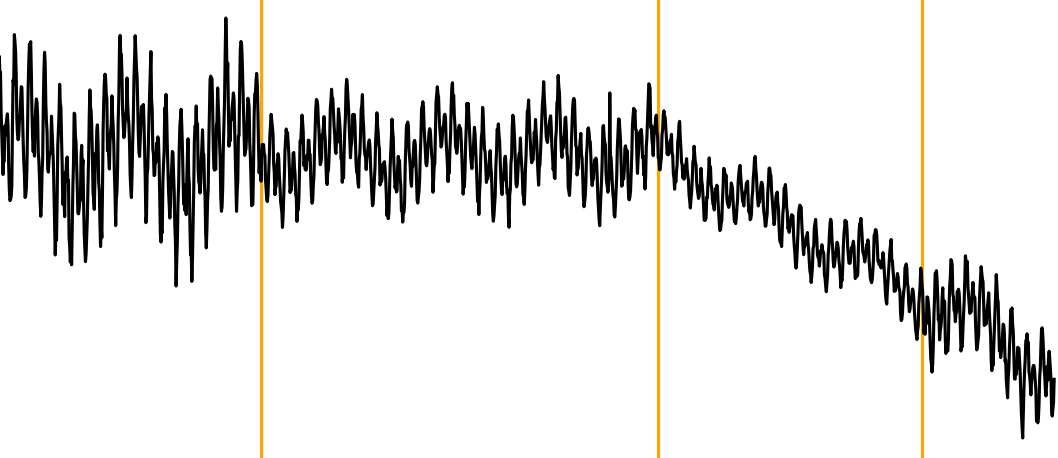}}
    \caption{Four random time series from the Yahoo! Benchmark dataset. The yellow vertical lines correspond to the change points. The spatial anomalies are not highlighted as they are not the focus of the experiment. }
    \label{fig:yahoo_example}
\end{figure}


\subsubsection{Yahoo! Benchmark Evaluation}
To compare the ability of each method to detect change points, we set three different detection error margins
of 24, 50, and 75 samples. If the difference between the actual and estimated change points were less than the specified margin, it was considered to be a true positive. 
Table \ref{tab:compare_baseline} shows the highest F1-score and corresponding window size for each method.

The Yahoo! Benchmark dataset is well-cited (according to \cite{wu2020current}) and one of the more complex datasets for temporal anomaly detection given the anomalies are mostly based upon changes in the seasonality, trend and noise. Based on the results reported in Table \ref{tab:compare_baseline}, our proposed method \TSCP\, strongly outperforms each of the other baseline methods.

Although all of the baselines are state-of-the-art methods for CPD, this dataset was shown to be challenging for them. Four randomly selected sequences of this dataset are illustrated in Figure \ref{fig:yahoo_example}. $FLOSS$ and $aHSIC$ were able to detect changes in temporal shape patterns, however, they could not distinguish the change points associated with subtle statistical differences. $RuLSIF$ estimate change points based upon the difference in the ratio of the distributions of adjacent time intervals. It was clear for some of the change points of the sequences in Figure \ref{fig:yahoo_example} that adjacent segments were similarly distributed and only exhibited clear changes in their temporal shape.  

\subsubsection{USC-HAD Dataset Evaluation}
Given the sampling rate for this dataset is 100Hz, the maximum error margins for which a change point estimate was considered to be a true positive was 1, 2, and 4 seconds. We investigate different values of the kernel bandwidth for \textit{RuLSIF} and different kernel sizes (20, 40, and 50) for \textit{aHSIC}. Different window size were investigated for \textit{FLOSS, ESPRESSO, KL-CPD}, and \TSCP\ as they were varied between 100, 200, and 400 samples. We also used different learning rates for \textit{KL-CPD} and \TSCP\ of $3\times10^{-4}$ and  $1\times10^{-4}$, respectively.

As shown in Table \ref{tab:compare_baseline}, our proposed method outperformed the other baselines across each of the error margins. \TSCP\ is the only method that delivers a high F1-score for the smallest error (100 samples) meaning it can reliably detect change points within one second of its occurrence.

\begin{figure}[!htbp]
    \centering
    \includegraphics[width=0.9\linewidth]{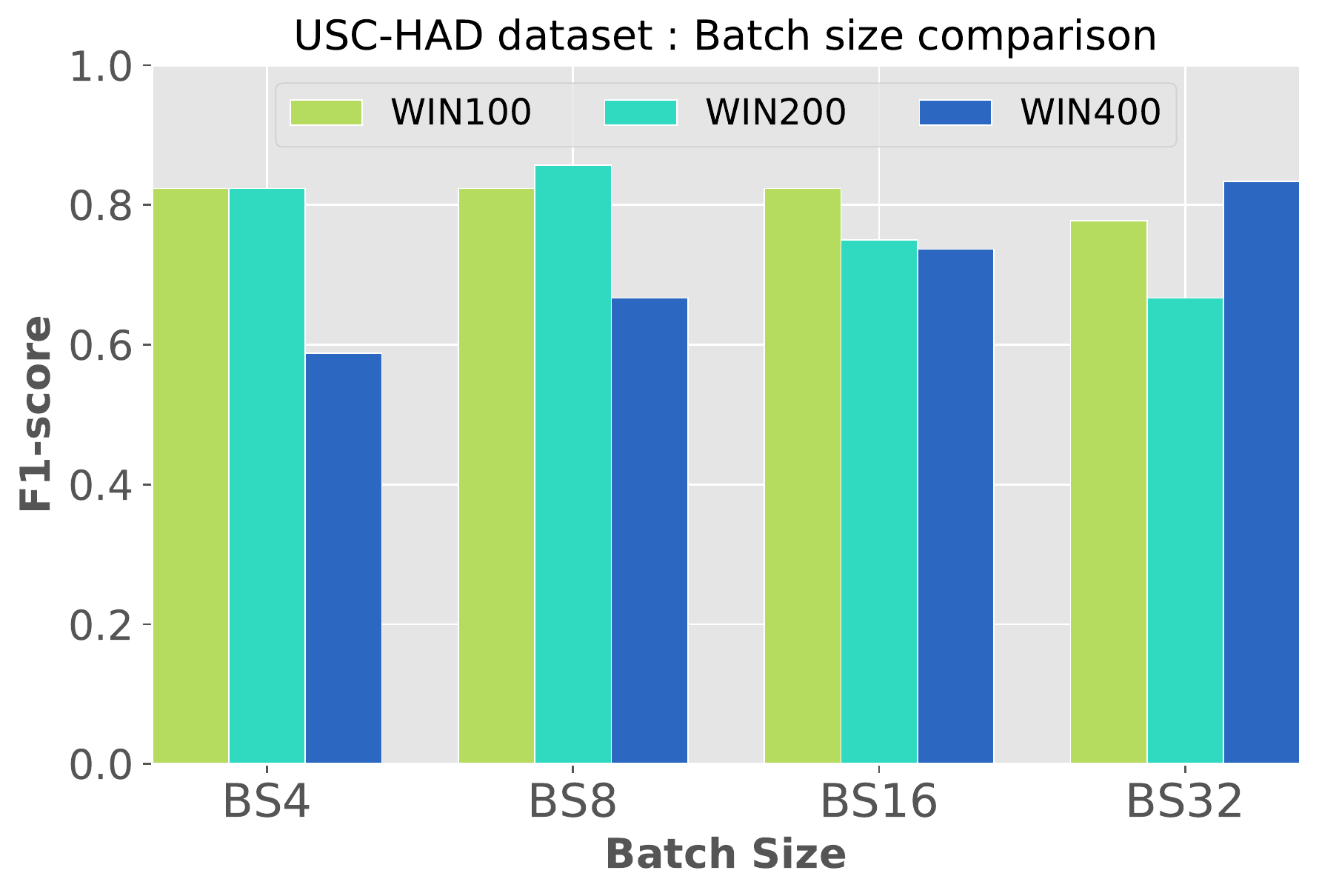}
    \caption{Comparing the effect of batch size across different window sizes for USC-HAD dataset.}
    \label{fig:usc_batch}
\end{figure}

Similarly to the Yahoo! Benchmark dataset, we compare the effect of batch size across the different window sizes for \TSCP\ in Figure \ref{fig:usc_batch}. It was shown that larger batch sizes offered a superior detection performance across the longer windows. The shorter batch sizes, however, were shown to offer superior detection performance across the smaller windows.

\subsubsection{HASC dataset Evaluation}
\label{sec:hasc}
The HASC dataset was found to be the most challenging dataset for \TSCP\ and the other baseline methods. Although \TSCP\ achieves the second highest performance level with respect to each of the different
window sizes, it still achieves the highest average F1-score with a \textbf{19.2\%, 54.8\%, 10.1\%, 11.1\%}, and \textbf{3.8\%} improvement over \textit{FLOSS, aHSIC, RulSIF, ESPRESSO,} and \textit{KL-CPD}, respectively.

The HASC dataset is the smallest in size (around 39K samples) but contains the largest number of change points (65 change points in total). Consequently, the relatively high density of change points means there there is a greater likelihood to generate positive sample pairs that encompass change points. Such false positive sample pairs will degrade model training. Since the model is self-supervised, ground truth labels cannot be used to rectify any such errors with positive sample pairs. Consequently, to address this problem, we suggest to enhance the model by injecting a light negative mining. We could also generate more positive sample pairs through augmentation.

\subsubsection{Discussion}
We showed that \TSCP\ was able to outperform non deep learning based methods, \textit{FLOSS, aHSIC, RulSIF} and \textit{ESPRESSO}, by \textbf{104.3\%, 91.0\%, 68.9\%, 53.3\%} and \textbf{24.9\%} improvement with respect to the F1-score averaged over all of the datasets. In addition, \TSCP\ showed a \textbf{17.0\%} improvement in F1-score over \textit{KL-CDP}, which is the most recent and competitive deep-learning-based change point detection method.

Since the baseline CPD methods exploit abrupt changes in one particular property of the time series, they do not effectively generalise to different types of datasets. For example, the hybrid \textit{ESPRESSO} method performs well on the USC-HAD dataset, given its change points are commonly associated with abrupt changes in both its temporal shape and statistical properties. But neither \textit{ESPRESSO} nor the other non deep learning approaches were as effective in estimating change points associated with the Yahoo!Benchmark given its change points were composed of more subtle and slowly evolving transitions in properties. In contrast, we showed that our proposed \TSCP\ method achieved either the first or second highest result across each of the datasets. \TSCP\ showed a significant improvement over the five baselines for the Yahoo! and USC-HAD datasets, whilst its average performance across the different window sizes was superior to each baseline with the HASC dataset.  

In future work, we will investigate using augmentation and negative mining batch construction to address the problem of high frequency change points that present themselves in some of the datasets discussed in section \ref{sec:hasc}. 

Finally, \TSCP\ employs a compact structure with a shared representation to encode its history and future windows. This enables faster training convergence compared to its other deep learning counterpart, \textit{KL-CPD}. 
Furthermore, once the representation model is trained, CPD is very simple to implement given it only involves a comparison between the learnt representations of the history and future windows. Consequently it has the potential to be implemented for online operation on low resource devices. The baseline methods (other than the \textit{FLOSS} method which has introduced a streaming version) cannot be applied online as they need to consider a reasonably large batch of data to capture repeated patterns or to optimise the entropy-based loss function.


%


\section{Conclusion}
We propose a novel \textbf{self-supervised} CPD method, \TSCP\, for time series. \TSCP\ learns an embedded representation  predict a future interval of a times series from historical samples. Change points are detected at the times in which the embedded are relatively high. Our proposed method is the first CPD method that employs contrastive learning to extract a compact and informative representation vector for every frame and estimate the change point based on the agreement between the learnt representation of subsequent frames.

We evaluated the ability of \TSCP\ in detecting change points against six other well-known state-of-the-art methods across three datasets. We have shown that our proposed method significantly outperform other baselines in two dataset and reaches a comparable score for the other dataset. Although the pre-trained \TSCP\ can detect changes in online applications, we aim to expand this method in our future work to continuously learn changes, anomalies and drifts in data.

\begin{acks}
We would like to acknowledge the support from CSIRO Data61 Scholarship program (Grant number 500588),  RMIT Research International Tuition Fee Scholarship (RRITFS), and Australian Research Council (ARC) Discovery Project \textit{DP190101485}.
\end{acks}

\bibliographystyle{ACM-Reference-Format}
\bibliography{main}


\begin{thebibliography}{57}


\ifx \showCODEN    \undefined \def \showCODEN     #1{\unskip}     \fi
\ifx \showDOI      \undefined \def \showDOI       #1{#1}\fi
\ifx \showISBNx    \undefined \def \showISBNx     #1{\unskip}     \fi
\ifx \showISBNxiii \undefined \def \showISBNxiii  #1{\unskip}     \fi
\ifx \showISSN     \undefined \def \showISSN      #1{\unskip}     \fi
\ifx \showLCCN     \undefined \def \showLCCN      #1{\unskip}     \fi
\ifx \shownote     \undefined \def \shownote      #1{#1}          \fi
\ifx \showarticletitle \undefined \def \showarticletitle #1{#1}   \fi
\ifx \showURL      \undefined \def \showURL       {\relax}        \fi
\providecommand\bibfield[2]{#2}
\providecommand\bibinfo[2]{#2}
\providecommand\natexlab[1]{#1}
\providecommand\showeprint[2][]{arXiv:#2}

\bibitem[\protect\citeauthoryear{Aakur and Sarkar}{Aakur and Sarkar}{2019}]%
        {aakur2019perceptual}
\bibfield{author}{\bibinfo{person}{Sathyanarayanan~N Aakur} {and}
  \bibinfo{person}{Sudeep Sarkar}.} \bibinfo{year}{2019}\natexlab{}.
\newblock \showarticletitle{A Perceptual Prediction Framework for Self
  Supervised Event Segmentation}. In \bibinfo{booktitle}{\emph{Proceedings of
  the IEEE Conference on Computer Vision and Pattern Recognition}}.
  \bibinfo{pages}{1197--1206}.
\newblock


\bibitem[\protect\citeauthoryear{Aminikhanghahi and Cook}{Aminikhanghahi and
  Cook}{2017}]%
        {Aminikhanghahi2017}
\bibfield{author}{\bibinfo{person}{Samaneh Aminikhanghahi} {and}
  \bibinfo{person}{Diane~J. Cook}.} \bibinfo{year}{2017}\natexlab{}.
\newblock \showarticletitle{A Survey of Methods for Time Series Change Point
  Detection}.
\newblock \bibinfo{journal}{\emph{Knowledge and information systems}}
  \bibinfo{volume}{51}, \bibinfo{number}{2} (\bibinfo{date}{01 May}
  \bibinfo{year}{2017}), \bibinfo{pages}{339--367}.
\newblock
\showISSN{0219-3116}


\bibitem[\protect\citeauthoryear{Aminikhanghahi and Cook}{Aminikhanghahi and
  Cook}{2019}]%
        {aminikhanghahi2019enhancing}
\bibfield{author}{\bibinfo{person}{Samaneh Aminikhanghahi} {and}
  \bibinfo{person}{Diane~J Cook}.} \bibinfo{year}{2019}\natexlab{}.
\newblock \showarticletitle{Enhancing Activity Recognition Using CPD-based
  Activity Segmentation}.
\newblock \bibinfo{journal}{\emph{Pervasive and Mobile Computing}}
  \bibinfo{volume}{53} (\bibinfo{year}{2019}).
\newblock


\bibitem[\protect\citeauthoryear{Basseville and Nikiforov}{Basseville and
  Nikiforov}{1993}]%
        {Basseville1993cdetection}
\bibfield{author}{\bibinfo{person}{Michelle Basseville} {and}
  \bibinfo{person}{Igor~V Nikiforov}.} \bibinfo{year}{1993}\natexlab{}.
\newblock \bibinfo{booktitle}{\emph{Detection of abrupt changes: theory and
  application}}.
\newblock \bibinfo{publisher}{Prentice Hall}.
\newblock


\bibitem[\protect\citeauthoryear{Chamroukhi, Mohammed, Trabelsi, Oukhellou, and
  Amirat}{Chamroukhi et~al\mbox{.}}{2013}]%
        {chamroukhi2013joint}
\bibfield{author}{\bibinfo{person}{Faicel Chamroukhi}, \bibinfo{person}{Samer
  Mohammed}, \bibinfo{person}{Dorra Trabelsi}, \bibinfo{person}{Latifa
  Oukhellou}, {and} \bibinfo{person}{Yacine Amirat}.}
  \bibinfo{year}{2013}\natexlab{}.
\newblock \showarticletitle{Joint Segmentation of Multivariate Time Series with
  Hidden Process Regression for Human Activity Recognition}.
\newblock \bibinfo{journal}{\emph{Neurocomputing}}  \bibinfo{volume}{120}
  (\bibinfo{year}{2013}), \bibinfo{pages}{633--644}.
\newblock


\bibitem[\protect\citeauthoryear{Chang, Li, Yang, and P{\'o}czos}{Chang
  et~al\mbox{.}}{2019}]%
        {chang2019kernel}
\bibfield{author}{\bibinfo{person}{Wei-Cheng Chang},
  \bibinfo{person}{Chun-Liang Li}, \bibinfo{person}{Yiming Yang}, {and}
  \bibinfo{person}{Barnab{\'a}s P{\'o}czos}.} \bibinfo{year}{2019}\natexlab{}.
\newblock \showarticletitle{Kernel change-point detection with auxiliary deep
  generative models}.
\newblock \bibinfo{journal}{\emph{arXiv preprint arXiv:1901.06077}}
  (\bibinfo{year}{2019}).
\newblock


\bibitem[\protect\citeauthoryear{Chavarriaga, Sagha, Calatroni, Digumarti,
  Tr{\"o}ster, Mill{\'a}n, and Roggen}{Chavarriaga et~al\mbox{.}}{2013}]%
        {chavarriaga2013opportunity}
\bibfield{author}{\bibinfo{person}{Ricardo Chavarriaga}, \bibinfo{person}{Hesam
  Sagha}, \bibinfo{person}{Alberto Calatroni}, \bibinfo{person}{Sundara~Tejaswi
  Digumarti}, \bibinfo{person}{Gerhard Tr{\"o}ster}, \bibinfo{person}{Jos{\'e}
  del~R Mill{\'a}n}, {and} \bibinfo{person}{Daniel Roggen}.}
  \bibinfo{year}{2013}\natexlab{}.
\newblock \showarticletitle{The Opportunity challenge: A benchmark Database for
  On-Body Sensor-based Activity Recognition}.
\newblock \bibinfo{journal}{\emph{Pattern Recognition Letters}}
  \bibinfo{volume}{34}, \bibinfo{number}{15} (\bibinfo{year}{2013}),
  \bibinfo{pages}{2033--2042}.
\newblock


\bibitem[\protect\citeauthoryear{Chen, Kornblith, Norouzi, and Hinton}{Chen
  et~al\mbox{.}}{2020}]%
        {chen2020simclr}
\bibfield{author}{\bibinfo{person}{Ting Chen}, \bibinfo{person}{Simon
  Kornblith}, \bibinfo{person}{Mohammad Norouzi}, {and}
  \bibinfo{person}{Geoffrey Hinton}.} \bibinfo{year}{2020}\natexlab{}.
\newblock \showarticletitle{A Simple Framework for Contrastive Learning of
  Visual Representations}.
\newblock \bibinfo{journal}{\emph{ICML}} (\bibinfo{year}{2020}).
\newblock


\bibitem[\protect\citeauthoryear{Chopra, Hadsell, and LeCun}{Chopra
  et~al\mbox{.}}{2005}]%
        {chopra2005contlosslearning}
\bibfield{author}{\bibinfo{person}{Sumit Chopra}, \bibinfo{person}{Raia
  Hadsell}, {and} \bibinfo{person}{Yann LeCun}.}
  \bibinfo{year}{2005}\natexlab{}.
\newblock \showarticletitle{Learning a similarity metric discriminatively, with
  application to face verification}. In \bibinfo{booktitle}{\emph{2005 IEEE
  Computer Society Conference on Computer Vision and Pattern Recognition
  (CVPR'05)}}, Vol.~\bibinfo{volume}{1}. IEEE, \bibinfo{pages}{539--546}.
\newblock


\bibitem[\protect\citeauthoryear{Dataset}{Dataset}{[n.d.]}]%
        {yahoodataset}
\bibfield{author}{\bibinfo{person}{Yahoo Research~Webscope Dataset}.}
  \bibinfo{year}{[n.d.]}\natexlab{}.
\newblock \showarticletitle{“S5 - A Labeled Anomaly Detection Dataset,
  version 1.0}.
\newblock  (\bibinfo{year}{[n.\,d.]}).
\newblock
\urldef\tempurl%
\url{https://webscope.sandbox.yahoo.com/}
\showURL{%
\tempurl}


\bibitem[\protect\citeauthoryear{De~Ryck, De~Vos, and Bertrand}{De~Ryck
  et~al\mbox{.}}{2020}]%
        {de2020change}
\bibfield{author}{\bibinfo{person}{Tim De~Ryck}, \bibinfo{person}{Maarten
  De~Vos}, {and} \bibinfo{person}{Alexander Bertrand}.}
  \bibinfo{year}{2020}\natexlab{}.
\newblock \showarticletitle{Change Point Detection in Time Series Data using
  Autoencoders with a Time-Invariant Representation}.
\newblock \bibinfo{journal}{\emph{arXiv preprint arXiv:2008.09524}}
  (\bibinfo{year}{2020}).
\newblock


\bibitem[\protect\citeauthoryear{Deldari, Liono, Salim, and Smith}{Deldari
  et~al\mbox{.}}{2019}]%
        {deldari2016inferring}
\bibfield{author}{\bibinfo{person}{Shohreh Deldari}, \bibinfo{person}{Jonathan
  Liono}, \bibinfo{person}{Flora~D Salim}, {and} \bibinfo{person}{Daniel~V
  Smith}.} \bibinfo{year}{2019}\natexlab{}.
\newblock \showarticletitle{Inferring Work Routines and Behavior Deviations
  with Life-logging Sensor Data}. In \bibinfo{booktitle}{\emph{Proceedings of
  ACM International Conference on Web Search and Data Mining (WSDM) workshop on
  Task Intelligence (TI@WSDM) (2019)}}. ACM.
\newblock


\bibitem[\protect\citeauthoryear{Deldari, Smith, Sadri, and Salim}{Deldari
  et~al\mbox{.}}{2020}]%
        {deldari2020Espresso}
\bibfield{author}{\bibinfo{person}{Shohreh Deldari}, \bibinfo{person}{Daniel~V.
  Smith}, \bibinfo{person}{Amin Sadri}, {and} \bibinfo{person}{Flora Salim}.}
  \bibinfo{year}{2020}\natexlab{}.
\newblock \showarticletitle{ESPRESSO: Entropy and ShaPe AwaRe TimE-Series
  SegmentatiOn for Processing Heterogeneous Sensor Data}.
\newblock \bibinfo{journal}{\emph{Proc. ACM Interact. Mob. Wearable Ubiquitous
  Technol.}} \bibinfo{volume}{4}, \bibinfo{number}{3}, Article
  \bibinfo{articleno}{77} (\bibinfo{date}{Sept.} \bibinfo{year}{2020}),
  \bibinfo{numpages}{24}~pages.
\newblock
\urldef\tempurl%
\url{https://doi.org/10.1145/3411832}
\showDOI{\tempurl}


\bibitem[\protect\citeauthoryear{Ding and Xu}{Ding and Xu}{2018}]%
        {ding2018weakly}
\bibfield{author}{\bibinfo{person}{Li Ding} {and} \bibinfo{person}{Chenliang
  Xu}.} \bibinfo{year}{2018}\natexlab{}.
\newblock \showarticletitle{Weakly-supervised action segmentation with
  iterative soft boundary assignment}. In \bibinfo{booktitle}{\emph{Proceedings
  of the IEEE Conference on Computer Vision and Pattern Recognition}}.
  \bibinfo{pages}{6508--6516}.
\newblock


\bibitem[\protect\citeauthoryear{Duan, Chen, Lu, and Zhou}{Duan
  et~al\mbox{.}}{2019}]%
        {duan2019deep}
\bibfield{author}{\bibinfo{person}{Yueqi Duan}, \bibinfo{person}{Lei Chen},
  \bibinfo{person}{Jiwen Lu}, {and} \bibinfo{person}{Jie Zhou}.}
  \bibinfo{year}{2019}\natexlab{}.
\newblock \showarticletitle{Deep embedding learning with discriminative
  sampling policy}. In \bibinfo{booktitle}{\emph{Proceedings of the IEEE
  Conference on Computer Vision and Pattern Recognition}}.
  \bibinfo{pages}{4964--4973}.
\newblock


\bibitem[\protect\citeauthoryear{Farha and Gall}{Farha and Gall}{2019}]%
        {farha2019ms}
\bibfield{author}{\bibinfo{person}{Yazan~Abu Farha} {and}
  \bibinfo{person}{Jurgen Gall}.} \bibinfo{year}{2019}\natexlab{}.
\newblock \showarticletitle{Ms-tcn: Multi-stage temporal convolutional network
  for action segmentation}. In \bibinfo{booktitle}{\emph{Proceedings of the
  IEEE Conference on Computer Vision and Pattern Recognition}}.
  \bibinfo{pages}{3575--3584}.
\newblock


\bibitem[\protect\citeauthoryear{Franceschi, Dieuleveut, and Jaggi}{Franceschi
  et~al\mbox{.}}{2019}]%
        {franceschi2019unsupervised}
\bibfield{author}{\bibinfo{person}{Jean-Yves Franceschi},
  \bibinfo{person}{Aymeric Dieuleveut}, {and} \bibinfo{person}{Martin Jaggi}.}
  \bibinfo{year}{2019}\natexlab{}.
\newblock \showarticletitle{Unsupervised scalable representation learning for
  multivariate time series}. In \bibinfo{booktitle}{\emph{Advances in Neural
  Information Processing Systems}}. \bibinfo{pages}{4650--4661}.
\newblock


\bibitem[\protect\citeauthoryear{Gharghabi, Yeh, Ding, Ding, Hibbing, LaMunion,
  Kaplan, Crouter, and Keogh}{Gharghabi et~al\mbox{.}}{2019}]%
        {gharghabi2018domain}
\bibfield{author}{\bibinfo{person}{Shaghayegh Gharghabi},
  \bibinfo{person}{Chin-Chia~Michael Yeh}, \bibinfo{person}{Yifei Ding},
  \bibinfo{person}{Wei Ding}, \bibinfo{person}{Paul Hibbing},
  \bibinfo{person}{Samuel LaMunion}, \bibinfo{person}{Andrew Kaplan},
  \bibinfo{person}{Scott~E Crouter}, {and} \bibinfo{person}{Eamonn Keogh}.}
  \bibinfo{year}{2019}\natexlab{}.
\newblock \showarticletitle{Domain Agnostic Online Semantic Segmentation for
  Multi-dimensional Time Series}.
\newblock \bibinfo{journal}{\emph{Data Mining and Knowledge Discovery}}
  \bibinfo{volume}{33}, \bibinfo{number}{1} (\bibinfo{year}{2019}),
  \bibinfo{pages}{96--130}.
\newblock


\bibitem[\protect\citeauthoryear{Gutmann and Hyv{\"a}rinen}{Gutmann and
  Hyv{\"a}rinen}{2010}]%
        {gutmann2010noise}
\bibfield{author}{\bibinfo{person}{Michael Gutmann} {and} \bibinfo{person}{Aapo
  Hyv{\"a}rinen}.} \bibinfo{year}{2010}\natexlab{}.
\newblock \showarticletitle{Noise-contrastive estimation: A new estimation
  principle for unnormalized statistical models}. In
  \bibinfo{booktitle}{\emph{Proceedings of the Thirteenth International
  Conference on Artificial Intelligence and Statistics}}.
  \bibinfo{pages}{297--304}.
\newblock


\bibitem[\protect\citeauthoryear{Hallac, Nystrup, and Boyd}{Hallac
  et~al\mbox{.}}{2019}]%
        {hallac2018greedy}
\bibfield{author}{\bibinfo{person}{David Hallac}, \bibinfo{person}{Peter
  Nystrup}, {and} \bibinfo{person}{Stephen Boyd}.}
  \bibinfo{year}{2019}\natexlab{}.
\newblock \showarticletitle{Greedy Gaussian Segmentation of Multivariate Time
  Series}.
\newblock \bibinfo{journal}{\emph{Advances in Data Analysis and
  Classification}} \bibinfo{volume}{13}, \bibinfo{number}{3}
  (\bibinfo{year}{2019}), \bibinfo{pages}{727--751}.
\newblock


\bibitem[\protect\citeauthoryear{H{\'e}naff, Srinivas, De~Fauw, Razavi,
  Doersch, Eslami, and Oord}{H{\'e}naff et~al\mbox{.}}{2020}]%
        {henaff2020data}
\bibfield{author}{\bibinfo{person}{Olivier~J H{\'e}naff},
  \bibinfo{person}{Aravind Srinivas}, \bibinfo{person}{Jeffrey De~Fauw},
  \bibinfo{person}{Ali Razavi}, \bibinfo{person}{Carl Doersch},
  \bibinfo{person}{SM Eslami}, {and} \bibinfo{person}{Aaron van~den Oord}.}
  \bibinfo{year}{2020}\natexlab{}.
\newblock \showarticletitle{Data-efficient image recognition with contrastive
  predictive coding}.
\newblock \bibinfo{journal}{\emph{ICML}} (\bibinfo{year}{2020}).
\newblock


\bibitem[\protect\citeauthoryear{Huang, Wang, Zhao, and Zhang}{Huang
  et~al\mbox{.}}{2019}]%
        {huang2019id}
\bibfield{author}{\bibinfo{person}{Anna Huang}, \bibinfo{person}{Dong Wang},
  \bibinfo{person}{Run Zhao}, {and} \bibinfo{person}{Qian Zhang}.}
  \bibinfo{year}{2019}\natexlab{}.
\newblock \showarticletitle{Au-Id: Automatic User Identification and
  Authentication Through the Motions Captured from Sequential Human Activities
  Using RFID}.
\newblock \bibinfo{journal}{\emph{Proceedings of the ACM on Interactive,
  Mobile, Wearable and Ubiquitous Technologies (IMWUT)}} \bibinfo{volume}{3},
  \bibinfo{number}{2} (\bibinfo{year}{2019}), \bibinfo{pages}{1--26}.
\newblock


\bibitem[\protect\citeauthoryear{Huang, Koh, Dobbie, and Pears}{Huang
  et~al\mbox{.}}{2014}]%
        {huang2014detecting}
\bibfield{author}{\bibinfo{person}{David Tse~Jung Huang},
  \bibinfo{person}{Yun~Sing Koh}, \bibinfo{person}{Gillian Dobbie}, {and}
  \bibinfo{person}{Russel Pears}.} \bibinfo{year}{2014}\natexlab{}.
\newblock \showarticletitle{Detecting Changes in Rare Patterns from Data
  Streams}. In \bibinfo{booktitle}{\emph{Pacific-Asia Conference on Knowledge
  Discovery and Data Mining (PAKDD)}}. Springer, \bibinfo{pages}{437--448}.
\newblock


\bibitem[\protect\citeauthoryear{Kawaguchi, Ogawa, Iwasaki, Kaji, Terada,
  Murao, Inoue, Kawahara, Sumi, and Nishio}{Kawaguchi et~al\mbox{.}}{2011a}]%
        {nobuo2011hasc}
\bibfield{author}{\bibinfo{person}{Nobuo Kawaguchi}, \bibinfo{person}{Nobuhiro
  Ogawa}, \bibinfo{person}{Yohei Iwasaki}, \bibinfo{person}{Katsuhiko Kaji},
  \bibinfo{person}{Tsutomu Terada}, \bibinfo{person}{Kazuya Murao},
  \bibinfo{person}{Sozo Inoue}, \bibinfo{person}{Yoshihiro Kawahara},
  \bibinfo{person}{Yasuyuki Sumi}, {and} \bibinfo{person}{Nobuhiko Nishio}.}
  \bibinfo{year}{2011}\natexlab{a}.
\newblock \showarticletitle{HASC Challenge: Gathering Large Scale Human
  Activity Corpus for the Real-World Activity Understandings}.
\newblock \bibinfo{journal}{\emph{ACM International Conference Proceeding
  Series}}, \bibinfo{pages}{27}.
\newblock
\urldef\tempurl%
\url{https://doi.org/10.1145/1959826.1959853}
\showDOI{\tempurl}


\bibitem[\protect\citeauthoryear{Kawaguchi, Yang, Yang, Ogawa, Iwasaki, Kaji,
  Terada, Murao, Inoue, Kawahara, et~al\mbox{.}}{Kawaguchi
  et~al\mbox{.}}{2011b}]%
        {kawaguchi2011hasccorpus}
\bibfield{author}{\bibinfo{person}{Nobuo Kawaguchi}, \bibinfo{person}{Ying
  Yang}, \bibinfo{person}{Tianhui Yang}, \bibinfo{person}{Nobuhiro Ogawa},
  \bibinfo{person}{Yohei Iwasaki}, \bibinfo{person}{Katsuhiko Kaji},
  \bibinfo{person}{Tsutomu Terada}, \bibinfo{person}{Kazuya Murao},
  \bibinfo{person}{Sozo Inoue}, \bibinfo{person}{Yoshihiro Kawahara},
  {et~al\mbox{.}}} \bibinfo{year}{2011}\natexlab{b}.
\newblock \showarticletitle{HASC2011corpus: towards the common ground of human
  activity recognition}. In \bibinfo{booktitle}{\emph{Proceedings of the 13th
  international conference on Ubiquitous computing}}.
  \bibinfo{pages}{571--572}.
\newblock


\bibitem[\protect\citeauthoryear{Kokkonen, Puuska, Alatalo, Heilimo, and
  M{\"a}kel{\"a}}{Kokkonen et~al\mbox{.}}{2019}]%
        {kokkonen2019network}
\bibfield{author}{\bibinfo{person}{Tero Kokkonen}, \bibinfo{person}{Samir
  Puuska}, \bibinfo{person}{Janne Alatalo}, \bibinfo{person}{Eppu Heilimo},
  {and} \bibinfo{person}{Antti M{\"a}kel{\"a}}.}
  \bibinfo{year}{2019}\natexlab{}.
\newblock \showarticletitle{Network anomaly detection based on wavenet}.
\newblock In \bibinfo{booktitle}{\emph{Internet of Things, Smart Spaces, and
  Next Generation Networks and Systems}}. \bibinfo{publisher}{Springer},
  \bibinfo{pages}{424--433}.
\newblock


\bibitem[\protect\citeauthoryear{Kurt, Y{\i}ld{\i}z, Ceritli, Sankur, and
  Cemgil}{Kurt et~al\mbox{.}}{2018}]%
        {kurt2018bayesian}
\bibfield{author}{\bibinfo{person}{Bar{\i}{\c{s}} Kurt},
  \bibinfo{person}{{\c{C}}a{\u{g}}atay Y{\i}ld{\i}z},
  \bibinfo{person}{Taha~Yusuf Ceritli}, \bibinfo{person}{B{\"u}lent Sankur},
  {and} \bibinfo{person}{Ali~Taylan Cemgil}.} \bibinfo{year}{2018}\natexlab{}.
\newblock \showarticletitle{A Bayesian change point model for detecting
  SIP-based DDoS attacks}.
\newblock \bibinfo{journal}{\emph{Digital Signal Processing}}
  \bibinfo{volume}{77} (\bibinfo{year}{2018}), \bibinfo{pages}{48--62}.
\newblock


\bibitem[\protect\citeauthoryear{Lam, Varona-Marin, Li, Fergenbaum, and
  Kuli{\'c}}{Lam et~al\mbox{.}}{2016}]%
        {lam2016automated}
\bibfield{author}{\bibinfo{person}{Agnes~WK Lam}, \bibinfo{person}{Danniel
  Varona-Marin}, \bibinfo{person}{Yeti Li}, \bibinfo{person}{Mitchell
  Fergenbaum}, {and} \bibinfo{person}{Dana Kuli{\'c}}.}
  \bibinfo{year}{2016}\natexlab{}.
\newblock \showarticletitle{Automated Rehabilitation System: Movement
  Measurement and Feedback for Patients and Physiotherapists in the
  Rehabilitation Clinic}.
\newblock \bibinfo{journal}{\emph{Human--Computer Interaction}}
  \bibinfo{volume}{31}, \bibinfo{number}{3-4} (\bibinfo{year}{2016}),
  \bibinfo{pages}{294--334}.
\newblock


\bibitem[\protect\citeauthoryear{Liono, Qin, and Salim}{Liono
  et~al\mbox{.}}{2016}]%
        {liono2016optimal}
\bibfield{author}{\bibinfo{person}{Jonathan Liono}, \bibinfo{person}{A~Kai
  Qin}, {and} \bibinfo{person}{Flora~D Salim}.}
  \bibinfo{year}{2016}\natexlab{}.
\newblock \showarticletitle{Optimal Time Window for Temporal Segmentation of
  Sensor Streams in multi-activity recognition}. In
  \bibinfo{booktitle}{\emph{Proceedings of the 13th International Conference on
  Mobile and Ubiquitous Systems: Computing, Networking and Services}}.
  \bibinfo{pages}{10--19}.
\newblock


\bibitem[\protect\citeauthoryear{Liu, Yamada, Collier, and Sugiyama}{Liu
  et~al\mbox{.}}{2013}]%
        {liu2013change}
\bibfield{author}{\bibinfo{person}{Song Liu}, \bibinfo{person}{Makoto Yamada},
  \bibinfo{person}{Nigel Collier}, {and} \bibinfo{person}{Masashi Sugiyama}.}
  \bibinfo{year}{2013}\natexlab{}.
\newblock \showarticletitle{Change-point Detection in Time-series Data by
  Relative Density-Ratio Estimation}.
\newblock \bibinfo{journal}{\emph{Neural Networks}}  \bibinfo{volume}{43}
  (\bibinfo{year}{2013}), \bibinfo{pages}{72--83}.
\newblock


\bibitem[\protect\citeauthoryear{Lu and Huang}{Lu and Huang}{2020}]%
        {lu2020segmentation}
\bibfield{author}{\bibinfo{person}{Shaowen Lu} {and} \bibinfo{person}{Shuyu
  Huang}.} \bibinfo{year}{2020}\natexlab{}.
\newblock \showarticletitle{Segmentation of Multivariate Industrial Time Series
  Data Based on Dynamic Latent Variable Predictability}.
\newblock \bibinfo{journal}{\emph{IEEE Access}}  \bibinfo{volume}{8}
  (\bibinfo{year}{2020}), \bibinfo{pages}{112092--112103}.
\newblock


\bibitem[\protect\citeauthoryear{Mnih and Kavukcuoglu}{Mnih and
  Kavukcuoglu}{2013}]%
        {mnih2013learning}
\bibfield{author}{\bibinfo{person}{Andriy Mnih} {and} \bibinfo{person}{Koray
  Kavukcuoglu}.} \bibinfo{year}{2013}\natexlab{}.
\newblock \showarticletitle{Learning word embeddings efficiently with
  noise-contrastive estimation}. In \bibinfo{booktitle}{\emph{Advances in
  neural information processing systems}}. \bibinfo{pages}{2265--2273}.
\newblock


\bibitem[\protect\citeauthoryear{Mounir, Gula, Theuerkauf, and Sarkar}{Mounir
  et~al\mbox{.}}{2020}]%
        {mounir2020temporal}
\bibfield{author}{\bibinfo{person}{Ramy Mounir}, \bibinfo{person}{Roman Gula},
  \bibinfo{person}{J{\"o}rn Theuerkauf}, {and} \bibinfo{person}{Sudeep
  Sarkar}.} \bibinfo{year}{2020}\natexlab{}.
\newblock \showarticletitle{Temporal Event Segmentation using Attention-based
  Perceptual Prediction Model for Continual Learning}.
\newblock \bibinfo{journal}{\emph{arXiv preprint arXiv:2005.02463}}
  (\bibinfo{year}{2020}).
\newblock


\bibitem[\protect\citeauthoryear{Oord, Dieleman, Zen, Simonyan, Vinyals,
  Graves, Kalchbrenner, Senior, and Kavukcuoglu}{Oord et~al\mbox{.}}{2016}]%
        {oord2016wavenet}
\bibfield{author}{\bibinfo{person}{Aaron van~den Oord}, \bibinfo{person}{Sander
  Dieleman}, \bibinfo{person}{Heiga Zen}, \bibinfo{person}{Karen Simonyan},
  \bibinfo{person}{Oriol Vinyals}, \bibinfo{person}{Alex Graves},
  \bibinfo{person}{Nal Kalchbrenner}, \bibinfo{person}{Andrew Senior}, {and}
  \bibinfo{person}{Koray Kavukcuoglu}.} \bibinfo{year}{2016}\natexlab{}.
\newblock \showarticletitle{Wavenet: A generative model for raw audio}.
\newblock \bibinfo{journal}{\emph{arXiv preprint arXiv:1609.03499}}
  (\bibinfo{year}{2016}).
\newblock


\bibitem[\protect\citeauthoryear{Oord, Li, and Vinyals}{Oord
  et~al\mbox{.}}{2018}]%
        {oord2018representation}
\bibfield{author}{\bibinfo{person}{Aaron van~den Oord}, \bibinfo{person}{Yazhe
  Li}, {and} \bibinfo{person}{Oriol Vinyals}.} \bibinfo{year}{2018}\natexlab{}.
\newblock \showarticletitle{Representation learning with contrastive predictive
  coding}.
\newblock \bibinfo{journal}{\emph{arXiv preprint arXiv:1807.03748}}
  (\bibinfo{year}{2018}).
\newblock


\bibitem[\protect\citeauthoryear{Sadri, Ren, and Salim}{Sadri
  et~al\mbox{.}}{2017}]%
        {sadri2017information}
\bibfield{author}{\bibinfo{person}{Amin Sadri}, \bibinfo{person}{Yongli Ren},
  {and} \bibinfo{person}{Flora~D Salim}.} \bibinfo{year}{2017}\natexlab{}.
\newblock \showarticletitle{Information Gain-based Metric for Recognizing
  Transitions in Human Activities}.
\newblock \bibinfo{journal}{\emph{Pervasive and Mobile Computing}}
  \bibinfo{volume}{38} (\bibinfo{year}{2017}), \bibinfo{pages}{92--109}.
\newblock


\bibitem[\protect\citeauthoryear{Sadri, Salim, Ren, Shao, Krumm, and
  Mascolo}{Sadri et~al\mbox{.}}{2018}]%
        {Sadri2018tajectory}
\bibfield{author}{\bibinfo{person}{Amin Sadri}, \bibinfo{person}{Flora~D
  Salim}, \bibinfo{person}{Yongli Ren}, \bibinfo{person}{Wei Shao},
  \bibinfo{person}{John~C Krumm}, {and} \bibinfo{person}{Cecilia Mascolo}.}
  \bibinfo{year}{2018}\natexlab{}.
\newblock \showarticletitle{What Will You Do for the Rest of the Day? an
  approach to continuous trajectory prediction}.
\newblock \bibinfo{journal}{\emph{Proceedings of the ACM on Interactive,
  Mobile, Wearable and Ubiquitous Technologies (IMWUT)}} \bibinfo{volume}{2},
  \bibinfo{number}{4} (\bibinfo{year}{2018}), \bibinfo{pages}{1--26}.
\newblock


\bibitem[\protect\citeauthoryear{Saeed, Grangier, and Zeghidour}{Saeed
  et~al\mbox{.}}{2020}]%
        {saeed2020contrastive}
\bibfield{author}{\bibinfo{person}{Aaqib Saeed}, \bibinfo{person}{David
  Grangier}, {and} \bibinfo{person}{Neil Zeghidour}.}
  \bibinfo{year}{2020}\natexlab{}.
\newblock \showarticletitle{Contrastive Learning of General-Purpose Audio
  Representations}.
\newblock \bibinfo{journal}{\emph{arXiv preprint arXiv:2010.10915}}
  (\bibinfo{year}{2020}).
\newblock


\bibitem[\protect\citeauthoryear{Saeed, Ozcelebi, and Lukkien}{Saeed
  et~al\mbox{.}}{2019}]%
        {saeed2019multi}
\bibfield{author}{\bibinfo{person}{Aaqib Saeed}, \bibinfo{person}{Tanir
  Ozcelebi}, {and} \bibinfo{person}{Johan Lukkien}.}
  \bibinfo{year}{2019}\natexlab{}.
\newblock \showarticletitle{Multi-task self-supervised learning for human
  activity detection}.
\newblock \bibinfo{journal}{\emph{Proceedings of the ACM on Interactive,
  Mobile, Wearable and Ubiquitous Technologies}} \bibinfo{volume}{3},
  \bibinfo{number}{2} (\bibinfo{year}{2019}), \bibinfo{pages}{1--30}.
\newblock


\bibitem[\protect\citeauthoryear{{Saeed}, {Salim}, {Ozcelebi}, and
  {Lukkien}}{{Saeed} et~al\mbox{.}}{2020}]%
        {saeed2020iot}
\bibfield{author}{\bibinfo{person}{A. {Saeed}}, \bibinfo{person}{F.~D.
  {Salim}}, \bibinfo{person}{T. {Ozcelebi}}, {and} \bibinfo{person}{J.
  {Lukkien}}.} \bibinfo{year}{2020}\natexlab{}.
\newblock \showarticletitle{Federated Self-Supervised Learning of Multi-Sensor
  Representations for Embedded Intelligence}.
\newblock \bibinfo{journal}{\emph{IEEE Internet of Things Journal}}
  (\bibinfo{year}{2020}), \bibinfo{pages}{1--1}.
\newblock


\bibitem[\protect\citeauthoryear{Schroff, Kalenichenko, and Philbin}{Schroff
  et~al\mbox{.}}{2015}]%
        {schroff2015facenet}
\bibfield{author}{\bibinfo{person}{Florian Schroff}, \bibinfo{person}{Dmitry
  Kalenichenko}, {and} \bibinfo{person}{James Philbin}.}
  \bibinfo{year}{2015}\natexlab{}.
\newblock \showarticletitle{Facenet: A unified embedding for face recognition
  and clustering}. In \bibinfo{booktitle}{\emph{Proceedings of the IEEE
  conference on computer vision and pattern recognition}}.
  \bibinfo{pages}{815--823}.
\newblock


\bibitem[\protect\citeauthoryear{Shoaib, Bosch, Incel, Scholten, and
  Havinga}{Shoaib et~al\mbox{.}}{2016}]%
        {shoaib2016complex}
\bibfield{author}{\bibinfo{person}{Muhammad Shoaib}, \bibinfo{person}{Stephan
  Bosch}, \bibinfo{person}{Ozlem~Durmaz Incel}, \bibinfo{person}{Hans
  Scholten}, {and} \bibinfo{person}{Paul~JM Havinga}.}
  \bibinfo{year}{2016}\natexlab{}.
\newblock \showarticletitle{Complex Human Activity Recognition Using Smartphone
  and Wrist-Worn Motion Sensors}.
\newblock \bibinfo{journal}{\emph{Sensors}} \bibinfo{volume}{16},
  \bibinfo{number}{4} (\bibinfo{year}{2016}), \bibinfo{pages}{426}.
\newblock


\bibitem[\protect\citeauthoryear{Simo-Serra, Trulls, Ferraz, Kokkinos, Fua, and
  Moreno-Noguer}{Simo-Serra et~al\mbox{.}}{2015}]%
        {simo2015discriminative}
\bibfield{author}{\bibinfo{person}{Edgar Simo-Serra}, \bibinfo{person}{Eduard
  Trulls}, \bibinfo{person}{Luis Ferraz}, \bibinfo{person}{Iasonas Kokkinos},
  \bibinfo{person}{Pascal Fua}, {and} \bibinfo{person}{Francesc
  Moreno-Noguer}.} \bibinfo{year}{2015}\natexlab{}.
\newblock \showarticletitle{Discriminative learning of deep convolutional
  feature point descriptors}. In \bibinfo{booktitle}{\emph{Proceedings of the
  IEEE International Conference on Computer Vision}}.
  \bibinfo{pages}{118--126}.
\newblock


\bibitem[\protect\citeauthoryear{Sohn}{Sohn}{2016}]%
        {sohn2016improved}
\bibfield{author}{\bibinfo{person}{Kihyuk Sohn}.}
  \bibinfo{year}{2016}\natexlab{}.
\newblock \showarticletitle{Improved deep metric learning with multi-class
  n-pair loss objective}. In \bibinfo{booktitle}{\emph{Advances in neural
  information processing systems}}. \bibinfo{pages}{1857--1865}.
\newblock


\bibitem[\protect\citeauthoryear{Wang, Hua, Kodirov, Hu, Garnier, and
  Robertson}{Wang et~al\mbox{.}}{2019}]%
        {wang2019ranked}
\bibfield{author}{\bibinfo{person}{Xinshao Wang}, \bibinfo{person}{Yang Hua},
  \bibinfo{person}{Elyor Kodirov}, \bibinfo{person}{Guosheng Hu},
  \bibinfo{person}{Romain Garnier}, {and} \bibinfo{person}{Neil~M Robertson}.}
  \bibinfo{year}{2019}\natexlab{}.
\newblock \showarticletitle{Ranked list loss for deep metric learning}. In
  \bibinfo{booktitle}{\emph{Proceedings of the IEEE Conference on Computer
  Vision and Pattern Recognition}}. \bibinfo{pages}{5207--5216}.
\newblock


\bibitem[\protect\citeauthoryear{Wang and Zheng}{Wang and Zheng}{2018}]%
        {wang2018modeling}
\bibfield{author}{\bibinfo{person}{Yanwen Wang} {and} \bibinfo{person}{Yuanqing
  Zheng}.} \bibinfo{year}{2018}\natexlab{}.
\newblock \showarticletitle{Modeling RFID Signal Reflection for Contact-free
  Activity Recognition}.
\newblock \bibinfo{journal}{\emph{Proceedings of the ACM on Interactive,
  Mobile, Wearable and Ubiquitous Technologies (IMWUT)}} \bibinfo{volume}{2},
  \bibinfo{number}{4}, Article \bibinfo{articleno}{193} (\bibinfo{year}{2018}),
  \bibinfo{numpages}{22}~pages.
\newblock
\urldef\tempurl%
\url{https://doi.org/10.1145/3287071}
\showDOI{\tempurl}


\bibitem[\protect\citeauthoryear{Wei, Wang, Nguyen, Zhang, Lin, Shen, Mech, and
  Samaras}{Wei et~al\mbox{.}}{2018}]%
        {wei2018sequence}
\bibfield{author}{\bibinfo{person}{Zijun Wei}, \bibinfo{person}{Boyu Wang},
  \bibinfo{person}{Minh~Hoai Nguyen}, \bibinfo{person}{Jianming Zhang},
  \bibinfo{person}{Zhe Lin}, \bibinfo{person}{Xiaohui Shen},
  \bibinfo{person}{Radom{\'\i}r Mech}, {and} \bibinfo{person}{Dimitris
  Samaras}.} \bibinfo{year}{2018}\natexlab{}.
\newblock \showarticletitle{Sequence-to-segment networks for segment
  detection}. In \bibinfo{booktitle}{\emph{Advances in Neural Information
  Processing Systems}}. \bibinfo{pages}{3507--3516}.
\newblock


\bibitem[\protect\citeauthoryear{Weinberger and Saul}{Weinberger and
  Saul}{2009}]%
        {weinberger2009distance}
\bibfield{author}{\bibinfo{person}{Kilian~Q Weinberger} {and}
  \bibinfo{person}{Lawrence~K Saul}.} \bibinfo{year}{2009}\natexlab{}.
\newblock \showarticletitle{Distance metric learning for large margin nearest
  neighbor classification.}
\newblock \bibinfo{journal}{\emph{Journal of Machine Learning Research}}
  \bibinfo{volume}{10}, \bibinfo{number}{2} (\bibinfo{year}{2009}).
\newblock


\bibitem[\protect\citeauthoryear{Wu, Manmatha, Smola, and Krahenbuhl}{Wu
  et~al\mbox{.}}{2017}]%
        {wu2017sampling}
\bibfield{author}{\bibinfo{person}{Chao-Yuan Wu}, \bibinfo{person}{R Manmatha},
  \bibinfo{person}{Alexander~J Smola}, {and} \bibinfo{person}{Philipp
  Krahenbuhl}.} \bibinfo{year}{2017}\natexlab{}.
\newblock \showarticletitle{Sampling matters in deep embedding learning}. In
  \bibinfo{booktitle}{\emph{Proceedings of the IEEE International Conference on
  Computer Vision}}. \bibinfo{pages}{2840--2848}.
\newblock


\bibitem[\protect\citeauthoryear{Wu and Keogh}{Wu and Keogh}{2020}]%
        {wu2020current}
\bibfield{author}{\bibinfo{person}{Renjie Wu} {and} \bibinfo{person}{Eamonn~J
  Keogh}.} \bibinfo{year}{2020}\natexlab{}.
\newblock \showarticletitle{Current Time Series Anomaly Detection Benchmarks
  are Flawed and are Creating the Illusion of Progress}.
\newblock \bibinfo{journal}{\emph{arXiv preprint arXiv:2009.13807}}
  (\bibinfo{year}{2020}).
\newblock


\bibitem[\protect\citeauthoryear{Xia, Korpela, Namioka, and Maekawa}{Xia
  et~al\mbox{.}}{2020}]%
        {xia2020robust}
\bibfield{author}{\bibinfo{person}{Qingxin Xia}, \bibinfo{person}{Joseph
  Korpela}, \bibinfo{person}{Yasuo Namioka}, {and} \bibinfo{person}{Takuya
  Maekawa}.} \bibinfo{year}{2020}\natexlab{}.
\newblock \showarticletitle{Robust Unsupervised Factory Activity Recognition
  with Body-worn Accelerometer Using Temporal Structure of Multiple Sensor Data
  Motifs}.
\newblock \bibinfo{journal}{\emph{Proceedings of the ACM on Interactive,
  Mobile, Wearable and Ubiquitous Technologies}} \bibinfo{volume}{4},
  \bibinfo{number}{3} (\bibinfo{year}{2020}), \bibinfo{pages}{1--30}.
\newblock


\bibitem[\protect\citeauthoryear{Yamada, Kimura, Naya, and Sawada}{Yamada
  et~al\mbox{.}}{2013}]%
        {Yamadaahsic}
\bibfield{author}{\bibinfo{person}{Makoto Yamada}, \bibinfo{person}{Akisato
  Kimura}, \bibinfo{person}{Futoshi Naya}, {and} \bibinfo{person}{Hiroshi
  Sawada}.} \bibinfo{year}{2013}\natexlab{}.
\newblock \showarticletitle{Change-point Detection with Feature Selection in
  High-dimensional Time-series Data}. In \bibinfo{booktitle}{\emph{Proc. of
  23th International Joint Conference on Artificial Intelligence (IJCAI)}}.
\newblock


\bibitem[\protect\citeauthoryear{Yamanishi and Takeuchi}{Yamanishi and
  Takeuchi}{2002}]%
        {yamanishi2002unifying}
\bibfield{author}{\bibinfo{person}{Kenji Yamanishi} {and}
  \bibinfo{person}{Jun-ichi Takeuchi}.} \bibinfo{year}{2002}\natexlab{}.
\newblock \showarticletitle{A unifying framework for detecting outliers and
  change points from non-stationary time series data}. In
  \bibinfo{booktitle}{\emph{Proceedings of the eighth ACM SIGKDD international
  conference on Knowledge discovery and data mining}}.
  \bibinfo{pages}{676--681}.
\newblock


\bibitem[\protect\citeauthoryear{Yoshinobu and Sugiyama}{Yoshinobu and
  Sugiyama}{2012}]%
        {Kawahara2012Seqeuntial}
\bibfield{author}{\bibinfo{person}{Kawahara Yoshinobu} {and}
  \bibinfo{person}{Masashi Sugiyama}.} \bibinfo{year}{2012}\natexlab{}.
\newblock \showarticletitle{Sequential Change-Point Detection Based on Direct
  Density-Ratio Estimation}.
\newblock \bibinfo{journal}{\emph{Statistical Analysis and Data Mining}}
  \bibinfo{volume}{5}, \bibinfo{number}{2} (\bibinfo{year}{2012}),
  \bibinfo{pages}{114--127}.
\newblock


\bibitem[\protect\citeauthoryear{Zameni, Sadri, Ghafoori, Moshtaghi, Salim,
  Leckie, and Ramamohanarao}{Zameni et~al\mbox{.}}{2019}]%
        {zameni2019unsupervised}
\bibfield{author}{\bibinfo{person}{Masoomeh Zameni}, \bibinfo{person}{Amin
  Sadri}, \bibinfo{person}{Zahra Ghafoori}, \bibinfo{person}{Masud Moshtaghi},
  \bibinfo{person}{Flora~D. Salim}, \bibinfo{person}{Christopher Leckie}, {and}
  \bibinfo{person}{Kotagiri Ramamohanarao}.} \bibinfo{year}{2019}\natexlab{}.
\newblock \showarticletitle{Unsupervised Online Change Point Detection in
  High-Dimensional Time Series}.
\newblock \bibinfo{journal}{\emph{Knowledge and Information Systems (KAIS)}}
  (\bibinfo{year}{2019}), \bibinfo{pages}{719--750}.
\newblock


\bibitem[\protect\citeauthoryear{Zhang and Sawchuk}{Zhang and Sawchuk}{2012}]%
        {zhang2012usc}
\bibfield{author}{\bibinfo{person}{Mi Zhang} {and} \bibinfo{person}{Alexander~A
  Sawchuk}.} \bibinfo{year}{2012}\natexlab{}.
\newblock \showarticletitle{USC-HAD: a Daily Activity Dataset for Ubiquitous
  Activity Recognition Using Wearable Sensors}. In
  \bibinfo{booktitle}{\emph{Proceedings of the 2012 ACM Conference on
  Ubiquitous Computing}} (Pittsburgh, Pennsylvania)
  \emph{(\bibinfo{series}{UbiComp ’12})}. \bibinfo{pages}{1036–1043}.
\newblock


\bibitem[\protect\citeauthoryear{Zhu, Zimmerman, Senobari, Yeh, Funning, Mueen,
  Brisk, and Keogh}{Zhu et~al\mbox{.}}{2016}]%
        {zhu2016matrix}
\bibfield{author}{\bibinfo{person}{Yan Zhu}, \bibinfo{person}{Zachary
  Zimmerman}, \bibinfo{person}{Nader~Shakibay Senobari},
  \bibinfo{person}{Chin-Chia~Michael Yeh}, \bibinfo{person}{Gareth Funning},
  \bibinfo{person}{Abdullah Mueen}, \bibinfo{person}{Philip Brisk}, {and}
  \bibinfo{person}{Eamonn Keogh}.} \bibinfo{year}{2016}\natexlab{}.
\newblock \showarticletitle{Matrix profile ii: Exploiting a novel algorithm and
  gpus to break the one hundred million barrier for time series motifs and
  joins}. In \bibinfo{booktitle}{\emph{2016 IEEE 16th international conference
  on data mining (ICDM)}}. IEEE, \bibinfo{pages}{739--748}.
\newblock


\end{thebibliography}

\end{document}